\documentclass[letterpaper]{article} % DO NOT CHANGE THIS
\usepackage{aaai2026}  % DO NOT CHANGE THIS
\usepackage{times}  % DO NOT CHANGE THIS
\usepackage{helvet}  % DO NOT CHANGE THIS
\usepackage{courier}  % DO NOT CHANGE THIS
\usepackage[hyphens]{url}  % DO NOT CHANGE THIS
\usepackage{graphicx} % DO NOT CHANGE THIS
\usepackage{natbib}  % DO NOT CHANGE THIS AND DO NOT ADD ANY OPTIONS TO IT
\usepackage{caption} % DO NOT CHANGE THIS AND DO NOT ADD ANY OPTIONS TO IT
\usepackage{algorithm}
\usepackage{algorithmic}
\usepackage{booktabs}
\usepackage{multirow}
\usepackage{xcolor}
\usepackage{amssymb}
\usepackage{amsmath}
\usepackage{array}
\usepackage{pifont}
\usepackage{newfloat}
\usepackage{listings}
\DeclareCaptionStyle{ruled}{labelfont=normalfont,labelsep=colon,strut=off} % DO NOT CHANGE THIS
\floatstyle{ruled}
\newfloat{listing}{tb}{lst}{}
\floatname{listing}{Listing}
\title{Predicting the Future by Retrieving the Past}
\author{
    Dazhao Du,
    Tao Han,
    Song Guo\thanks{Corresponding Author.}
}
\affiliations{
    Hong Kong University of Science and Technology\\
    \{dduab, thanad\}@connect.ust.hk, songguo@cse.ust.hk
}

\usepackage{bibentry}
\begin{document}

\maketitle

\begin{abstract}
Deep learning models such as MLP, Transformer, and TCN have achieved remarkable success in univariate time series forecasting, typically relying on sliding window samples from historical data for training. However, while these models implicitly compress historical information into their parameters during training, they are unable to explicitly and dynamically access this global knowledge during inference, relying only on the local context within the lookback window. This results in an underutilization of rich patterns from the global history. To bridge this gap, we propose \textbf{Predicting the Future by Retrieving the Past (PFRP)}, a novel approach that explicitly integrates global historical data to enhance forecasting accuracy. Specifically, we construct a \textbf{Global Memory Bank (GMB)} to effectively store and manage global historical patterns. A retrieval mechanism is then employed to extract similar patterns from the GMB, enabling the generation of global predictions. By adaptively combining these global predictions with the outputs of any local prediction model, PFRP produces more accurate and interpretable forecasts. Extensive experiments conducted on seven real-world datasets demonstrate that PFRP significantly enhances the average performance of advanced univariate forecasting models by 8.4\%. Codes can be found in \url{https://github.com/ddz16/PFRP}.  
\end{abstract}

% However, these models discard the historical data after training and rely solely on the local context within the lookback window for predictions, resulting in the underutilization of global historical information.

%     \link{Code}{https://anonymous.4open.science/r/PFRP-6278}

\section{Introduction}

\begin{figure}[t]
\centering
\includegraphics[width=0.97\columnwidth]{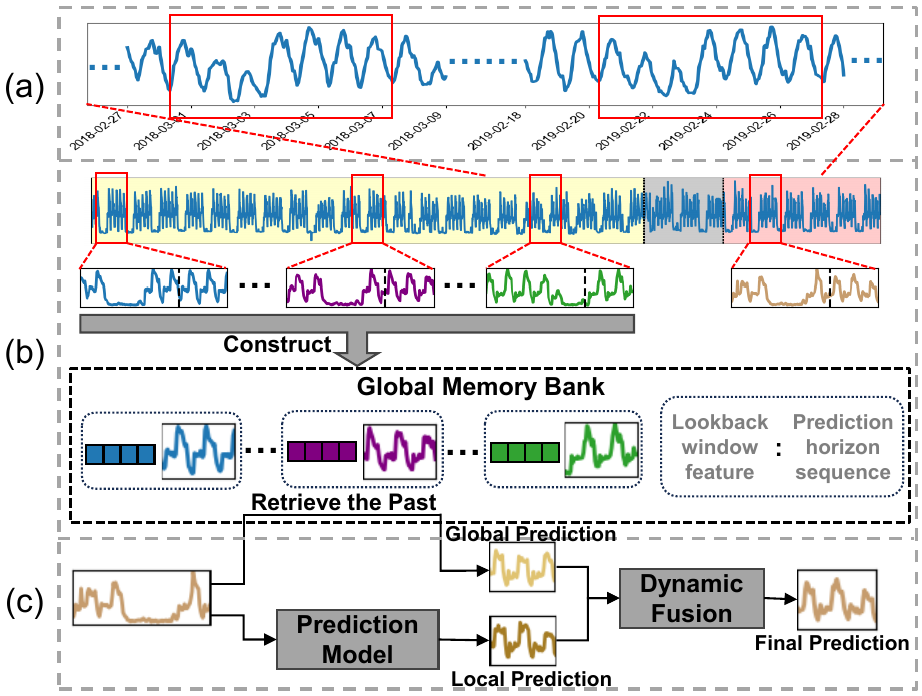}
\caption{(a) Time series often contain highly similar subsequences across different periods. (b) The GMB is constructed from historical sliding window samples, containing pairs of lookback window features and their corresponding prediction horizon sequences. (c) During inference, relevant patterns are retrieved from the GMB to generate global prediction, which are then dynamically fused with local prediction from any prediction model to yield the final result.}
\label{fig:intro}
\vspace{-0.2cm}
\end{figure}

Time series forecasting (TSF) has broad applications across various domains, including weather~\cite{fengwu}, finance~\cite{stock}, transportation~\cite{traffic}, and energy~\cite{electricity}. Recently, many advanced deep learning models have been proposed, such as Transformer-based~\cite{informer,timexer}, TCN-based~\cite{micn}, and MLP-based~\cite{dlinear} architectures. Despite their architectural differences, these models generally follow a common training and testing paradigm~\cite{lim2021time}. As illustrated in Figure~\ref{fig:intro}, an entire time series is typically partitioned into three intervals for training, validation, and test sets. A sliding window approach is then employed to create samples, each consisting of a lookback window sequence and a prediction horizon sequence. The training objective is to learn a mapping function from the lookback window sequence to the prediction horizon sequence. Upon completion of training, while the global historical information is implicitly compressed into the model's parameters, the original training data is typically discarded after training. Consequently, during inference, the trained model can only leverage the limited context within the current lookback window. These models are referred to as \textit{local prediction models}, as they lack the ability to explicitly reference specific, relevant patterns from the entire historical sequence to inform the prediction at hand.

% Upon completion of training, the historical information from the training set is assumed to be compressed into the model parameters, after which the training set is discarded. During inference, the trained model uses the current lookback window sequence to generate predictions. These models are referred to as \textit{local prediction models}, as they rely solely on the limited context provided by the lookback window.

However, we observe that time series often contain subsequences from different periods that exhibit remarkably similar patterns. For instance, in the household electricity consumption dataset~\cite{informer}, the consumption pattern from a week in 2019 closely resembles that of a week in 2018, as shown in Figure~\ref{fig:intro}(a). Both exhibit daily periodic fluctuations, with peaks gradually declining over the first three days and stabilizing at a high level for the subsequent four days. This observation suggests that TSF should not only rely solely on the local lookback window but also leverage the global historical sequence as a reference. Inspired by the retrieval-augmented generation technique in NLP~\cite{lewis2020retrieval,guu2020retrieval}, we propose a novel framework, Predicting the Future by Retrieving the Past (PFRP), which explicitly stores historical data and retrieves relevant sequences to enhance forecasting accuracy. PFRP addresses two key challenges: (1) \textit{how to effectively retrieve relevant historical sequences} and (2) \textit{how to integrate the retrieved sequences to improve current predictions}.

% chang2018memory
To address the first challenge, we construct a Global Memory Bank (GMB)~\cite{chang2018memory} to store the lookback window features and corresponding prediction horizon sequences from historical training samples, as shown in Figure~\ref{fig:intro}(b). The lookback window features serve as retrieval keys, while the prediction horizon sequences act as the associated values to be retrieved. During forecasting, we hypothesize that if the current lookback window sequence closely resembles a historical lookback window sequence, their respective prediction horizon sequences should also exhibit similar patterns. For instance, in Figure~\ref{fig:intro}(a), the first three days of two weeks are similar, and their subsequent four days also exhibit similar patterns. 

To address the second challenge, a naive approach is to directly copy the retrieved prediction horizon sequence as the current prediction result. However, this naive strategy performs suboptimally due to the inherent randomness and uncertainty in TSF. Furthermore, referencing multiple historical sequences is desirable to enhance robustness. To this end, we retrieve the top-k most similar lookback window features and compute a weighted combination of their corresponding prediction horizon sequences based on similarity scores. Additionally, we introduce a confidence gate and an output gate to dynamically adjust the weighting coefficients and modulate the scale and shift of the global prediction output. Although PFRP can independently generate global predictions, we find that dynamically fusing these global predictions with the local predictions of any other local prediction model yields superior prediction results.

Our contributions can be summarized as follows: (1) We propose the Global Memory Bank, a novel mechanism for explicitly storing and utilizing global historical data. (2) Building upon GMB, we introduce PFRP, a retrieval-based forecasting method that generates global predictions by retrieving and leveraging relevant historical patterns. (3) By seamlessly integrating global predictions with the outputs of any local prediction model, our model-agnostic approach significantly enhances univariate TSF performance.

\section{Related Work}
\subsection{Time Series Forecasting}
In the era of deep learning, numerous time series forecasting models have been proposed. From the perspective of model architectures, these models can be broadly categorized into RNN-based~\cite{deepar}, MLP-based~\cite{nbeats,timemixer}, CNN-based~\cite{timesnet,micn}, and Transformer-based~\cite{tft,preformer,fedformer} approaches. These methods focus on designing effective strategies for temporal modeling. Autoformer~\cite{autoformer} employs auto-correlation mechanisms, while DLinear~\cite{dlinear} utilizes simple linear mapping. From the data perspective, prediction models fall into two main categories: univariate and multivariate. Univariate models are designed to capture temporal patterns within a single variable, whereas multivariate models additionally address the inter-variable dependencies. For instance, iTransformer~\cite{itransformer} leverages attention to directly model inter-variable relationships. However, PatchTST~\cite{PatchTST} demonstrated that repeatedly applying univariate forecasting along the variable dimension can effectively achieve multivariate forecasting. Our method also focuses on univariate TSF. Furthermore, recent years have witnessed a surge in prediction models based on large language models~\cite{zhou2023one,jin2023time} and pre-trained foundation models~\cite{woo2024moirai,goswami2024moment}. Although a wide variety of models have been developed, most are essentially local prediction models, as they rely solely on limited lookback windows for forecasting, without fully utilizing the entire historical sequence. In contrast, our method integrates global historical data into the forecasting process.

\begin{table}[t]
\centering
\footnotesize
\setlength{\tabcolsep}{4pt}
\begin{tabular}{lccc}
\toprule
Method & Retrieval Criterion & Efficiency & Plug-and-Play\\
\midrule
RATD & Feature Similarity & \ding{55} & \ding{55} \\
TimeRAF & Feature Similarity & \ding{55} & \ding{55} \\
TimeRAG & DTW Distance & \ding{55} & \ding{55} \\
RAFT & Pearson Correlation & \ding{55}  & \ding{55} \\
\midrule
PFRP & Feature Similarity & \ding{51}  & \ding{51} \\
\bottomrule
\end{tabular}
\caption{Comparison between PFRP and existing RAG-based forecasting methods. Note that RATD, TimeRAF, and TimeRAG require diffusion models, time series foundation models, and LLMs respectively, resulting in significantly lower prediction efficiency. While RAFT requires retrieval across the entire training dataset, PFRP operates solely on a fixed-size memory bank, yielding higher retrieval efficiency.}
\label{tab:method_comparison}
\end{table}

\subsection{Retrieval-Augmented Generation}
In NLP, retrieval-augmented generation (RAG) combines pre-trained models with an external knowledge retrieval mechanism, allowing the model to dynamically access and integrate relevant information to enhance generation tasks~\cite{lewis2020retrieval,guu2020retrieval}. Recently, this technique has been extended to time series forecasting. RATD~\cite{liuretrieval} retrieves the most relevant historical time series to guide the denoising process in diffusion models. RAF~\cite{tire2024retrieval} and TimeRAF~\cite{zhang2024timeraf} leverage retrieval-augmented techniques to enhance the zero-shot forecasting capabilities of time series foundation models, while TimeRAG~\cite{yang2024timerag} focuses on improving LLM-based time series forecasting using retrieval-augmented methods. RAFT~\cite{raft} leverages retrieval from training data to augment the input. Table~\ref{tab:method_comparison} lists the comparison between PFRP and these approaches. Compared to them, PFRP offers a more efficient retrieval mechanism and prediction generation strategy. Additionally, PFRP is model-agnostic and adaptable for enhancing any existing univariate forecasting model.

\section{Methodology}
\subsection{Definition and Overview}
For univariate time series forecasting, suppose the training set comprises \(N\) historical sliding-window samples, denoted as \(\{(x^{(1)},y^{(1)}),(x^{(2)},y^{(2)}),\cdots,(x^{(N)},y^{(N)})\}\), where \(x^{(i)}\in \mathbb{R}^L\) represents the lookback window sequence and \(y^{(i)}\in \mathbb R^H\) denotes the prediction horizon sequence. Typically, a local prediction model is trained using these samples to learn the mapping function between \(x^{(i)}\) and \(y^{(i)}\). During inference, the current lookback window sequence \(x\) is input into the trained model to generate the corresponding prediction \(y\). Consequently, the training set is discarded once the prediction model is trained. In contrast, our method retains part of the training set and explicitly leverages the rich historical information it contains.

\begin{figure}[t]
\centering
\includegraphics[width=0.96\columnwidth]{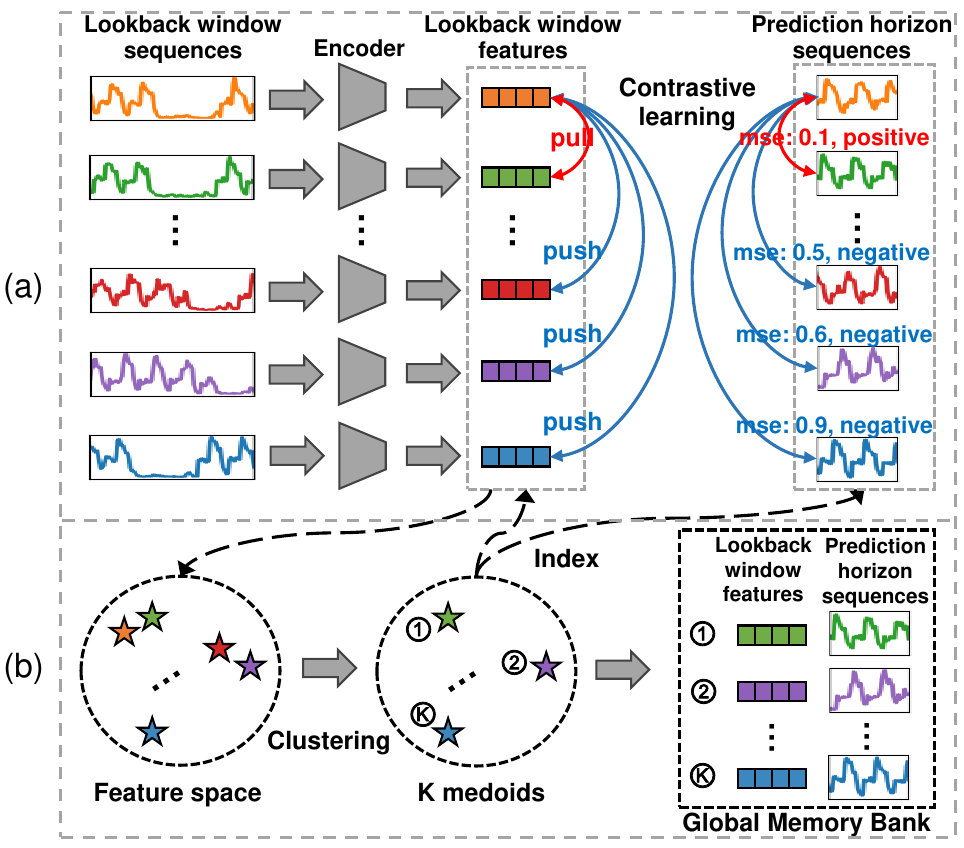}
\caption{Construction of GMB. (a) \textit{Predictive Contrastive Learning.} Positive sample pairs are identified as those whose prediction horizon sequences exhibit the lowest MSE. PCL aims to pull the encoded lookback window sequences of positive pairs closer in feature space. (b) \textit{K-medoids Clustering.} Retain only $K$ representative medoids in feature space to construct GMB, which stores both their lookback window features and corresponding prediction horizon sequences.}
\label{fig:GMB}
\end{figure}

\begin{figure*}[t]
\centering
\includegraphics[width=0.9\linewidth]{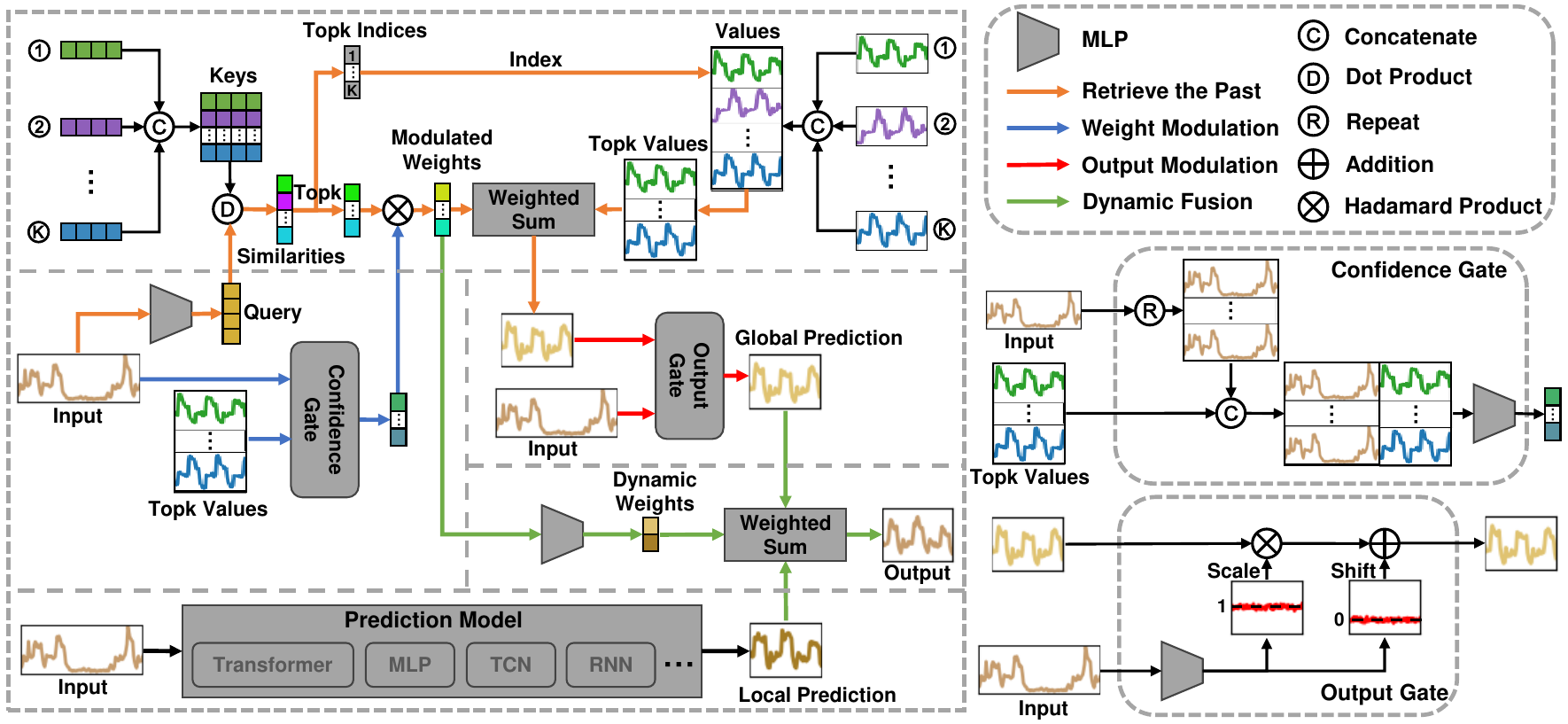}
\caption{Schematic diagram of PFRP. The diagram can be interpreted through the following key processes: (1) Orange arrows represent the retrieval process from GMB. (2) Blue arrows denote the weight modulation process. (3) Red arrows illustrate the output modulation process, which generates the global prediction. (4) The local prediction process is depicted at the bottom. (5) Green arrows indicate the fusion of global and local predictions based on dynamic weights. (6) The bottom-right section details the structures of the confidence gate and output gate.}
\label{fig:PFRP}
\end{figure*}

Our proposed method consists of two main stages. The first stage involves constructing a \textit{Global Memory Bank (GMB)} to store historical information. As illustrated in Figure~\ref{fig:GMB}, we introduce \textit{Predictive Contrastive Learning (PCL)} to train a encoder that encodes the lookback window sequences of all historical samples into high-level features. To reduce redundancy and improve retrieval efficiency, we apply \textit{K-medoids clustering} in the feature space, retaining only the cluster medoids. The second stage focuses on prediction through GMB retrieval, i.e., predicting the future by retrieving the past, as shown in Figure~\ref{fig:PFRP}. The global prediction retrieved from the GMB is dynamically fused with the local prediction produced by any prediction model to generate the final forecast. These two stages are detailed below.

\subsection{Global Memory Bank}

\subsubsection{Predictive Contrastive Learning}
To retrieve relevant historical samples based on the lookback window, some methods directly measure the similarity of lookback window sequences using DTW or MSE~\cite{yang2024timerag,zhang2024timeraf}. In contrast, we measure similarity at the feature level. To achieve this, we introduce contrastive learning~\cite{simclr} to train an MLP-based feature encoder for the lookback window sequences. We propose a new strategy for selecting positive and negative samples. Specifically, instead of selecting based on the MSE between lookback window sequences, we select them based on the MSE between their corresponding prediction horizon sequences. Intuitively, this training objective encourages lookback window sequences with more similar future to be closer in the feature space, which facilitates the retrieval of historical samples that are more helpful for the current prediction. We refer to this method as Predictive Contrastive Learning (PCL), as shown in Figure~\ref{fig:GMB}(a). Specifically, for $i$-th sample $(x^{(i)},y^{(i)})$ in a training batch \(\{(x^{(1)},y^{(1)}),\cdots,(x^{(B)},y^{(B)})\}\), where $B$ is the batch size, its positive sample index $i^+$ is:
\begin{equation}
i^+ = \arg\min_{\substack{1 \leq j \leq B, j \ne i}} \| y_i-y_j \|_2^2.
\end{equation}
Other samples in the same batch can be regarded as negative samples. We pass the lookback window sequence $x^{(i)}$ of each sample through the encoder to obtain the feature $\epsilon^{(i)}$. The objective function for PCL is then defined as:
\begin{equation}
\mathcal{L}_{pcl} = -\frac{1}{B}\sum_{i=1}^B\log{\frac{\exp{(\epsilon^{(i)}\cdot\epsilon^{(i^+)}/\tau)}}{\sum_{j=1,j\neq i}^B\exp{(\epsilon^{(i)}\cdot\epsilon^{(j)}/\tau})}},
\end{equation}
where $\tau$ is the temperature. 

\subsubsection{K-medoids Clustering} 
Using the time series encoder (a MLP) trained by PCL to encode the lookback window sequences of all training samples, we can obtain a new set \(\{(\epsilon^{(1)},y^{(1)}),(\epsilon^{(2)},y^{(2)}),\cdots,(\epsilon^{(N)},y^{(N)})\}\), where each sample is represented as a pair of a lookback window feature $\epsilon^{(i)}$ and its corresponding prediction horizon sequence $y^{(i)}$. To reduce redundancy and improve retrieval efficiency, we apply K-medoids clustering~\cite{Kmedoids} to the lookback window features, retaining only the samples corresponding to the $K$ cluster medoids, as shown in Figure~\ref{fig:GMB}(b). A key advantage of K-medoids over alternatives like K-means is its use of actual historical samples as cluster centroids rather than synthetic averages. This exemplar-based approach is essential for our task, as it ensures the patterns stored in our GMB represent authentic and coherent historical sequences. These selected samples are stored in the GMB, which can be formalized as \(\{(\epsilon^{(1)},y^{(1)}),(\epsilon^{(2)},y^{(2)}),\cdots,(\epsilon^{(K)},y^{(K)})\}\).

\subsection{Predicting the Future by Retrieving the Past}

\subsubsection{Retrieving from Global Memory Bank}
Firstly, the current lookback window sequence \(x\) is encoded by the encoder into a feature vector \(\epsilon\). This vector serves as the query and is used to compute cosine similarity with the $K$ historical lookback window features (keys) stored in the GMB:
\begin{equation}
    w^{(i)}=\epsilon\cdot\epsilon^{(i)},\quad i=\{1,\cdots,K\}.
\end{equation}
Assume the indices of the top-$k$ biggest similarities are given by $\{a_1,\cdots,a_k\}$. The top-$k$ biggest similarities are $\{w^{(a_1)},\cdots,w^{(a_k)}\}$, and the corresponding prediction horizon sequences (values) are $\{y^{(a_1)},\cdots,y^{(a_k)}\}$.

% Therefore, the weights corresponding to these $K$ keys are $\{w^{(1)},w^{(2)},\cdots,w^{(K)}\}$. 

The naive implementation of PFRP can be adopting an attention-like~\cite{vaswani2017attention} operation, where the query-key similarities are regarded as weights to compute a weighted sum of the values to generate the output. To adaptively regulate the entire process and enhance model capacity, we further introduce two learnable components: (i) a \textit{confidence gate} that adaptively adjusts attention weights, and (ii) an \textit{output gate} that modulates the global prediction.

\subsubsection{Confidence Gate}
Historical lookback window sequences that are more similar to $x$ do not necessarily indicate that their corresponding prediction horizon sequences are more likely to represent the future of $x$, nor do they warrant larger weights. Therefore, we design a confidence gate to modulate the weights. If the retrieved historical prediction horizon sequence $y^{(a_i)}$ better matches the current lookback window sequence $x$, then the complete sequence formed by concatenating them over time is more likely to exist, suggesting that the corresponding $y^{(a_i)}$ is more likely to represent the true future of $x$. To achieve this, we concatenate the top-$k$ retrieved values (i.e., the $k$ historical prediction horizon sequences) with $x$ to form $k$ complete sequences $\{[x;y^{(a_1)}],[x;y^{(a_2)}],\cdots,[x;y^{(a_k)}]\}$. Next, we use an MLP with a sigmoid activation function to output the existence probability for each of these $k$ complete sequences:
\begin{equation}
p_i = \textup{Sigmoid}(\textup{MLP}([x;y^{(a_i)}])),\quad p_i\in(0,1).
\end{equation}
These probabilities are then used to directly modulate the weights by multiplying them with the original top-$k$ weights $\{w^{(a_1)},w^{(a_2)},\cdots,w^{(a_k)}\}$:
\begin{equation}
\text{\small$\overline{w}^{(a_1)},\cdots,\overline{w}^{(a_k)}$} = \text{\small$\textup{Softmax}(w^{(a_1)} \cdot p_1,\cdots,w^{(a_k)} \cdot p_k)$}.
\end{equation}

\subsubsection{Output Gate}
The retrieved values are aggregated using a weighted sum based on the modulated weights to obtain the initial global prediction $\overline{y}_1$:
\begin{equation}
\overline{y}_1 = \sum_{i=1}^{k}\overline{w}^{(a_i)} \cdot y^{(a_i)}.
\end{equation}
However, the future sequence to be predicted may exhibit a pattern similar to a certain past sequence but might not align perfectly in terms of scale and shift. To address this, we employ an output gate to dynamically modulate the output based on the current lookback window sequence $x$. The output gate refines the initial global prediction $\overline{y}_1$ via learnable scale and shift. Specifically, $x$ is fed into an MLP that outputs two sequences, $\alpha\in \mathbb R^H$ and $\beta\in \mathbb R^H$. $\alpha$ represents the scale of the prediction horizon sequence and is initialized to all ones, while $\beta$ represents the shift and is initialized to all zeros. Then the global prediction $y_1$ is formulated as:
\begin{equation}
y_1 = \alpha \cdot \overline{y}_1 + \beta.
\end{equation}

\begin{table*}[t]
  \centering
  \footnotesize
  \setlength{\tabcolsep}{2.pt}
    \begin{tabular}{c|cccc|cccc|cccc|cccc}
    \toprule
    Models & \multicolumn{2}{c}{SparseTSF} & \multicolumn{2}{c|}{+PFRP} & \multicolumn{2}{c}{DLinear} & \multicolumn{2}{c|}{+PFRP} & \multicolumn{2}{c}{PatchTST} & \multicolumn{2}{c|}{+PFRP} & \multicolumn{2}{c}{TimesNet} & \multicolumn{2}{c}{+PFRP} \\
    \midrule
    Metrics & MSE   & MAE   & MSE   & MAE   & MSE   & MAE   & MSE   & MAE   & MSE   & MAE   & MSE   & MAE   & MSE   & MAE   & MSE   & MAE \\
    \midrule
    Traffic & 0.2404 & 0.3082 & \textbf{0.1919} & \textbf{0.2710} & 0.2778 & 0.3660 & \textbf{0.1793} & \textbf{0.2720} & 0.1797 & 0.2680 & \textbf{0.1712} & \textbf{0.2416} & 0.2165 & 0.3065 & \textbf{0.1799} & \textbf{0.2631} \\
    \midrule
    Electricity & 0.4968 & 0.5207 & \textbf{0.3561} & \textbf{0.4214} & 0.3951 & 0.4579 & \textbf{0.3666} & \textbf{0.4351} & 0.4200 & 0.4620 & \textbf{0.3869} & \textbf{0.4456} & 0.4264 & 0.4644 & \textbf{0.3950} & \textbf{0.4517} \\
    \midrule
    Weather & 0.6452 & 0.5724 & \textbf{0.6365} & \textbf{0.5659} & 0.7250 & 0.5961 & \textbf{0.6751} & \textbf{0.5735} & 0.6542 & 0.5719 & \textbf{0.6426} & \textbf{0.5663} & 0.6808 & 0.5853 & \textbf{0.6382} & \textbf{0.5676} \\
    \midrule
    ETTh1 & 0.0841 & 0.2247 & \textbf{0.0766} & \textbf{0.2141} & 0.1160 & 0.2594 & \textbf{0.1122} & \textbf{0.2564} & 0.0802 & 0.2180 & \textbf{0.0769} & \textbf{0.2145} & 0.0785 & 0.2172 & \textbf{0.0766} & \textbf{0.2147} \\
    \midrule
    ETTh2 & 0.2024 & 0.3537 & \textbf{0.1915} & \textbf{0.3437} & 0.2242 & 0.3694 & \textbf{0.2147} & \textbf{0.3621} & 0.2008 & 0.3531 & \textbf{0.1893} & \textbf{0.3411} & 0.1926 & 0.3459 & \textbf{0.1917} & \textbf{0.3442} \\
    \midrule
    ETTm1 & 0.0536 & 0.1752 & \textbf{0.0523} & \textbf{0.1710} & 0.0637 & 0.1838 & \textbf{0.0627} & \textbf{0.1828} & 0.0534 & 0.1735 & \textbf{0.0527} & \textbf{0.1719} & 0.0537 & 0.1744 & \textbf{0.0526} & \textbf{0.1717} \\
    \midrule
    ETTm2 & 0.1263 & 0.2684 & \textbf{0.1203} & \textbf{0.2570} & 0.1255 & 0.2622 & \textbf{0.1239} & \textbf{0.2596} & 0.1244 & 0.2621 & \textbf{0.1218} & \textbf{0.2585} & 0.1236 & 0.2606 & \textbf{0.1214} & \textbf{0.2584} \\
    \bottomrule
    \end{tabular}
 \caption{Univariate forecasting results for four baselines with and without PFRP. Lower metric values indicate better performance, and the better results are highlighted in bold. The lookback window length $L$ is set to 96. The results are averaged across all prediction horizons $H= \{96, 192, 336, 720\}$. See Table~\ref{tab:full_result} in the Appendix for the full results.}
  \label{tab:result}%
\end{table*}%

\subsubsection{Dynamic Fusion}
If no highly similar sequences exist in the historical data for the current lookback window, relying solely on GMB may lead to inaccurate predictions, and the model should reduce reliance on the global prediction. In such cases, the modulated weights $\{\overline{w}^{(a_1)},\overline{w}^{(a_2)},\cdots,\overline{w}^{(a_k)}\}$ tend to be relatively small. Therefore, we input the current lookback window sequence $x$ into a local prediction model to generate a local prediction $y_2$. By dynamically fusing the global prediction $y_1$ with the local prediction $y_2$, we achieve a more accurate forecast. Specifically, the modulated weights $\{\overline{w}^{(a_1)},\overline{w}^{(a_2)},\cdots,\overline{w}^{(a_k)}\}$ which reflect the importance of global versus local predictions are fed into an MLP followed by a Softmax activation function to dynamically compute the fusion weights:
\begin{equation}
w_1, w_2 = \textup{Softmax}(\textup{MLP}(\overline{w}^{(a_1)},\overline{w}^{(a_2)},\cdots,\overline{w}^{(a_k)})).
\end{equation}
Then the global prediction $y_1$ and local prediction $y_2$ are aggregated using a weighted sum based on the dynamic fusion weights $w_1, w_2$:
\begin{equation}
    y = w_1 \cdot y_1 + w_2 \cdot y_2,
\end{equation}
where $y$ is the final prediction result.

\section{Experiments}

\paragraph{Datasets}
We perform extensive experiments on seven datasets across three domains, including Traffic, Electricity, Weather, and four ETT datasets (ETTh1, ETTh2, ETTm1, ETTm2). While these datasets include multiple variables, our study focuses exclusively on univariate TSF. Accordingly, for each dataset, we consider only the time series of the last variable. Adhering to the standard forecasting setup~\cite{autoformer}, we fix the lookback window length $L$ at 96, with the prediction horizon $H$ varying across $\{96,192,336,720\}$. Forecasting performance is evaluated using the MSE and MAE metrics.

\paragraph{Baselines}
We select four SOTA local prediction models as the main baselines: two MLP-based models (DLinear~\cite{dlinear} and SparseTSF~\cite{sparsetsf}), one Transformer-based model (PatchTST~\cite{PatchTST}), and one CNN-based model (TimesNet~\cite{timesnet}). We further investigate whether PFRP can enhance the predictive performance of state-of-the-art large time-series models, including the LLM-based models (TimeCMA~\cite{timecma}) and time series foundation models (Moirai~\cite{woo2024moirai} and Sundial~\cite{sundial}). For these large models with strong zero-shot capabilities, we freeze their pretrained parameters and finetune only the PFRP-specific parameters. Additionally, we also compare PFRP with two RAG-based prediction methods: RATD~\cite{liuretrieval} and RAFT~\cite{raft}.

% The first three models are specifically tailored for univariate TSF, while the last one is versatile enough to be directly applied to univariate TSF as well.

\begin{figure*}[ht]
\centering
\includegraphics[width=0.97\linewidth]{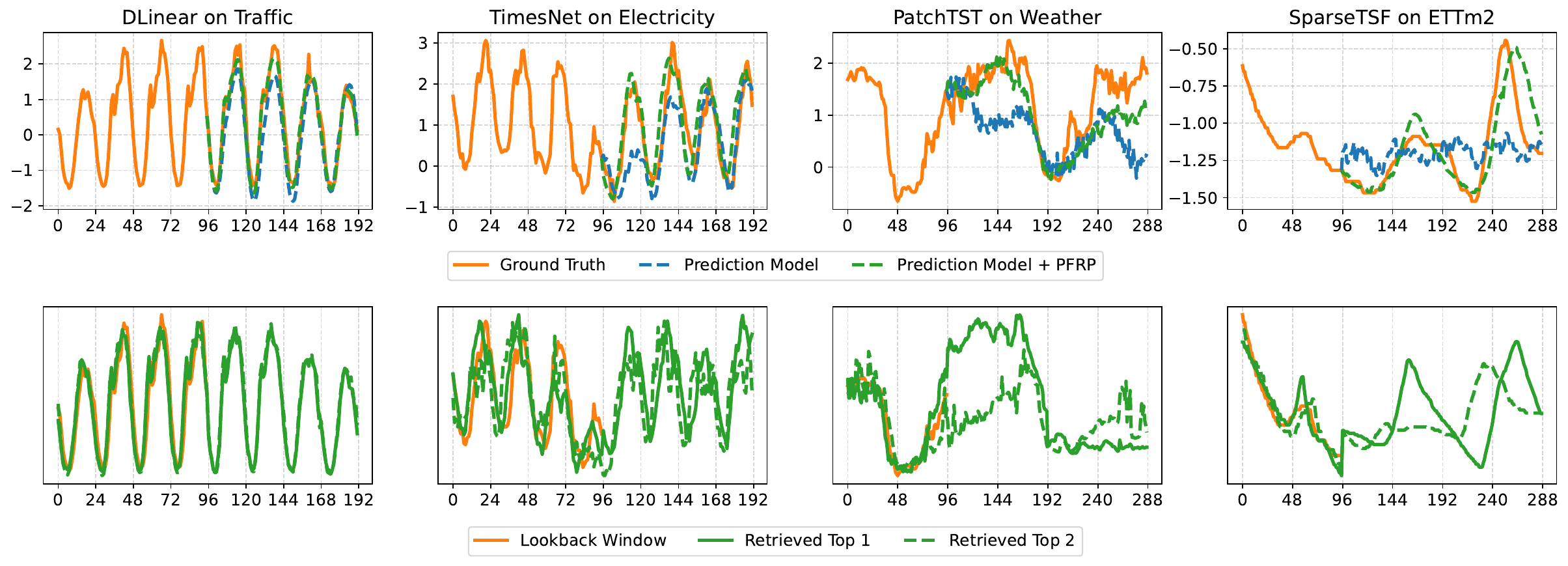}
\caption{\textit{Top:} Visualization of the prediction results for four baseline models, both with and without PFRP. \textit{Bottom:} Visualization of the top 2 most relevant historical sequences retrieved by PFRP.}
\label{fig:result}
\end{figure*}

\begin{figure}[t]
\centering
\includegraphics[width=\columnwidth]{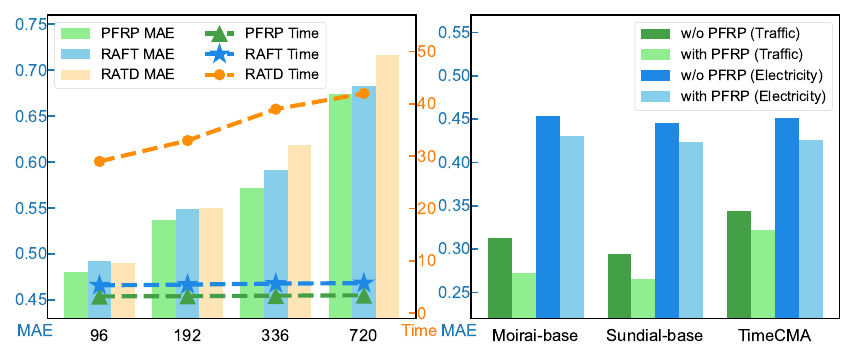}
\caption{\textit{Left:} Performance and efficiency comparison with two RAG-based prediction methods on the Weather dataset. \textit{Right:} Performance comparison of three large time-series models (with/without PFRP) on two datasets.}
\label{fig:morecomparisons}
\end{figure}

\paragraph{Implementation Details}
\textbf{First stage:} When training the lookback window encoder with PCL, we set the batch size to 256, the temperature to 0.05, and the learning rate to 0.001. To identify positive and negative samples within the training batch, we exclude those with significant temporal overlap ($>48$ same timestamps) with the anchor. For constructing the GMB, the lookback window length is set to 96, and the prediction horizon to 720. As a result, the GMB does not need to be reconstructed for each prediction horizon. For example, when the prediction horizon is 96/192/336, we simply extract the first 96/192/336 time steps from the stored prediction horizon sequence as the retrieved sequence. \textbf{Second stage:} When training four baseline models and our PFRP, we use the Adam optimizer~\cite{kingma2014adam} with an L2 loss function and an initial learning rate of 0.0001. For different local prediction models, we follow the batch size, training epochs, and hyperparameter settings from their official implementations. To ensure a fair comparison, we keep the model hyperparameters and training configurations unchanged, regardless of whether PFRP is applied. Each experiment is repeated three times with different random seeds, and we report the average results. All experiments are conducted on a single NVIDIA TITAN V GPU.

\begin{figure}[t]
\centering
\includegraphics[width=0.85\columnwidth]{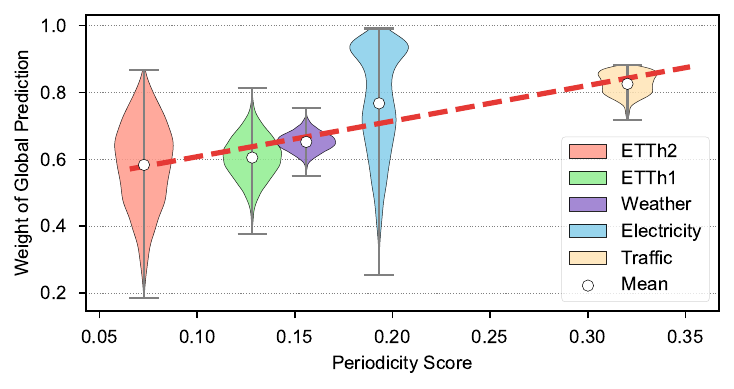}
\caption{The x-axis represents the datasets' periodicity scores, while the y-axis represents the weights $w_1$ of the global prediction from the Dynamic Fusion module for all test set samples.}
\label{fig:periodicity}
\end{figure}

\subsection{Main Result}
Table~\ref{tab:result} presents the forecasting results of baseline models and PFRP across seven datasets. We observe that PFRP consistently enhances the performance of these local prediction models. The most significant improvements are observed in DLinear and SparseTSF, two simple MLP-based models, with average gains of 7.1\% and 8.4\% across all datasets. For more complex models like PatchTST and TimesNet, the performance boost from PFRP is slightly smaller. On datasets with strong regularities, such as Traffic and Electricity, PFRP improves the forecasting accuracy of all local prediction models by an average of 17.4\% and 10.1\%, respectively. This suggests that in datasets with well-defined periodic patterns, similar sequences frequently recur, making our retrieval-based prediction approach particularly effective. Figure~\ref{fig:result} visualizes the predictions generated by the baselines and PFRP. Figure~\ref{fig:morecomparisons} (left) further compares PFRP (+SparseTSF) with two other RAG-based prediction methods. PFRP consistently outperforms both alternatives across varying prediction horizons. PFRP also exhibits superior test-time inference speed by retrieving solely from the GMB, unlike RAFT, which requires exhaustive traversal of the entire training set. Conversely, RATD shows the lowest efficiency due to its multi-step diffusion sampling. In Figure~\ref{fig:morecomparisons} (right), we find that incorporating large time series models (LLM-based models or foundation models) into our PFRP framework can also enhance performance.

\subsection{More Analysis}
% We conduct ablation experiments on the Traffic and Electricity datasets, and select SparseTSF as the default local prediction model.

\paragraph{Can PFRP effectively retrieve relevant sequences to improve prediction performance?} 
The bottom row in Figure~\ref{fig:result} visualizes the top two relevant historical sequences retrieved by PFRP. On the relatively simpler Traffic and Electricity datasets, the retrieved historical sequences compensate for peaks, refining the local predictions. On more challenging datasets such as Weather and ETT, while precise forecasting remains difficult, the retrieved historical sequences still help align predictions more closely with the actual future values. Besides, retrieved past temporal patterns provides interpretability to the current forecasting process.

\paragraph{Does PFRP demonstrate performance improvements on datasets with weaker periodicity?}
To quantify the degree of periodicity in time series datasets, we combine autocorrelation at predefined lags (e.g., daily, weekly) with the inverse of normalized entropy. The resulting score, ranging from 0 to 1, reflects the strength of the dataset's periodic structure. As Figure~\ref{fig:periodicity} illustrates, the Traffic and Electricity datasets exhibit the strongest periodicity, while the four ETT datasets show the weakest. Although improvements on the ETT datasets are less pronounced in Table~\ref{tab:result}, PFRP still achieves average gains of 3.4\%, 3.1\%, 1.6\%, and 2.2\%, respectively. We further observe that the global prediction weight ($w_1$) is directly proportional to the dataset's periodicity: for more periodic data, the final prediction relies more heavily on retrieval-based global prediction, shifting dependence away from the local prediction model's output.

\begin{table}[t]
  \centering
  \footnotesize
    \setlength{\tabcolsep}{2.5pt}{
    \begin{tabular}{c|c|cc|cc}
    \toprule
    \multicolumn{1}{r}{} &       & \multicolumn{2}{c|}{Traffic} & \multicolumn{2}{c}{Electricity} \\
    \midrule
    \multicolumn{1}{r}{} &       & MSE   & MAE   & MSE   & MAE \\
    \midrule
    \multirow{4}[2]{*}{\shortstack{Retrieval\\Criterion}} & Feature & \textbf{0.1919} & \textbf{0.2710} & \textbf{0.3561} & \textbf{0.4214} \\
          & MSE & 0.1928  & 0.2752  & 0.3716  & 0.4296  \\
          & DTW & 0.1922  & 0.2758  & 0.3659  & 0.4248  \\
          & PCC & 0.2020  & 0.2809  & 0.3789  & 0.4339  \\
    \bottomrule
    \multirow{3}[2]{*}{\shortstack{Encoder\\Type}} & MLP   & 0.1919  & 0.2710  & 0.3561  & 0.4214  \\
          & PatchTST & 0.1985  & 0.2760  & \textbf{0.3492} & \textbf{0.4175} \\
          & TimesNet & \textbf{0.1754} & \textbf{0.2599 } & 0.3754  & 0.4319  \\
    \midrule
    \multirow{3}[2]{*}{\shortstack{Training\\Strategy}} & PCL & \textbf{0.1919} & \textbf{0.2710} & \textbf{0.3561} & \textbf{0.4214} \\
          & CL & 0.2101  & 0.2828  & 0.3833  & 0.4350  \\
          & PL & 0.2250  & 0.3012  & 0.3762  & 0.4325  \\
    \midrule
    \end{tabular}}
    \caption{Ablations about GMB on two datasets.}
  \label{tab:abla_GMB}%
\end{table}%

% \paragraph{Does the feature similarity-based retrieval approach outperform retrieval methods based on other metrics?}

% \paragraph{Do encoders trained with predictive contrastive learning exhibit superior performance compared to those trained with either contrastive learning or predictive learning alone?}

\paragraph{Ablations about GMB}
We explore various aspects of the GMB, including retrieval criterion, encoder types, and encoder training strategies. (1) \textbf{Retrieval Criterion}: We use cosine similarity of encoded features as the retrieval criterion and compare it with non-feature-based approaches that directly measure raw lookback window sequence similarities. For instance, TimeRAG~\cite{yang2024timerag} employs DTW, while we also test MSE and the Pearson Correlation Coefficient (PCC). However, these methods prove less effective than our feature-based cosine similarity retrieval. (2) \textbf{Lookback Window Encoder}: We replace the default MLP encoder with two widely used time series encoders, PatchTST and TimesNet. Results show that the optimal encoder may vary by dataset and needs to be determined experimentally: TimesNet achieves the highest accuracy on the Traffic dataset, while PatchTST excels on Electricity. However, PFRP improves local prediction regardless of the encoder used. For efficiency in building the GMB, we select MLP as the default encoder. (3) \textbf{Encoder Training Strategy}: We compare three training strategies. Predictive Learning (PL) attaches a prediction head to the encoder and optimizes it with a standard forecasting task, as used in RATD~\cite{liuretrieval}. Contrastive Learning (CL) selects positive samples based on lookback window sequence similarity. In contrast, our Predictive Contrastive Learning (PCL) selects positives based on the similarity of their prediction horizon sequences. Compared to the other two methods, PCL aligns better with the retrieval-based forecasting objective, enabling it to identify historical samples with more similar future behaviors.

\begin{table}[t]
  \centering
  \footnotesize
  \setlength{\tabcolsep}{2.5pt}{
  \begin{tabular}{c|cc|cc}
\toprule      & \multicolumn{2}{c|}{Traffic} & \multicolumn{2}{c}{Electricity} \\
\midrule
       & MSE   & MAE   & MSE   & MAE \\
    \midrule
    SparseTSF & 0.2404  & 0.3082  & 0.4968  & 0.5207  \\
    \midrule
    SparseTSF+PFRP  & 0.1919  & 0.2710  & \textbf{0.3561} & \textbf{0.4214} \\
          w/o confidence gate & 0.2385  & 0.3058  & 0.3960  & 0.4486  \\
          w/o output gate & 0.2130  & 0.2989  & 0.5140  & 0.5296  \\
          w/o both gates & 0.2128  & 0.2987  & 0.5763  & 0.5643  \\
          w/o prediction model & \textbf{0.1686} & \textbf{0.2476} & 0.3952  & 0.4494  \\
    \bottomrule
    \end{tabular}}
    \caption{Ablations about PFRP on two datasets.}
  \label{tab:abla_PFRP}%
\end{table}%

\paragraph{Ablations about PFRP}
We conduct ablation experiments to investigate the impact of three learnable components in PFRP: the confidence gate, the output gate, and the local prediction model. The results in Table~\ref{tab:abla_PFRP} show that removing either of the two gates, or both, leads to a decline in forecasting performance. Furthermore, we observe that even when the local prediction model is removed, leaving only the global prediction results, our method still outperforms the baseline. On the Electricity dataset, dynamically combining global and local predictions yields better results. Conversely, on the Traffic dataset, relying solely on global predictions can achieve even better performance.

% \begin{figure}[t]
% \begin{center}
% \includegraphics[width=0.95\columnwidth]{figures/sensitivity.pdf}
% \caption{The impact of key hyperparameters $K$ and $k$ in PFRP, as well as the lookback window length $L$, on forecasting performance. We explore the impact of $L$ on performance both with and without the local prediction model. See Figure~\ref{fig:sensitivity1} in the Appendix for the results on other five datasets.}
% \label{fig:sensitivity}
% \end{center}
% \end{figure}

\paragraph{Efficiency Analysis}

\begin{figure}[t]
\centering
\includegraphics[width=0.85\columnwidth]{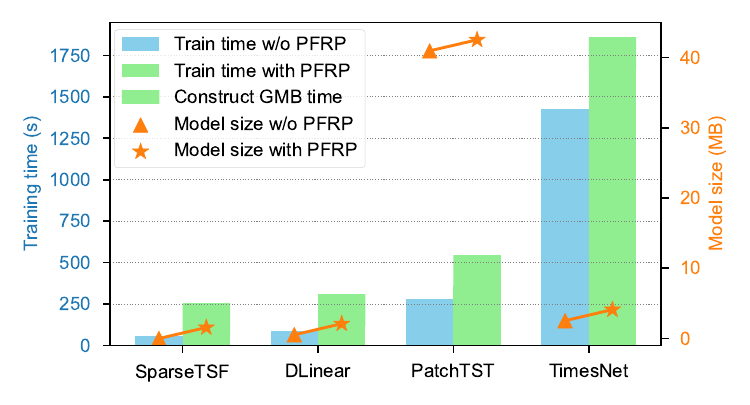}
\caption{Total consumed time and model sizes of four baseline models with and without PFRP.}
\label{fig:efficiency}
\end{figure}

Figure~\ref{fig:efficiency} compares the model size and training time of four baseline models with and without PFRP, using the Electricity dataset (prediction horizon is 720). For a fair comparison, all methods were trained for 10 epochs. While GMB construction adds a fixed 186 seconds (134 seconds for PCL, 52 seconds for K-medoids clustering), PFRP's overall impact on efficiency is minimal. It consistently increases model size by only 1.57 MB and causes only a slight rise in training duration, underscoring PFRP's practicality and efficiency.

More experiments and analysis are in the Appendix.

\section{Conclusion}

We introduce PFRP, a retrieval-enhanced univariate time series forecasting framework that explicitly incorporates historical patterns to enhance prediction accuracy. Unlike conventional local prediction models that rely solely on a fixed-length lookback window, our approach leverages a Global Memory Bank (GMB) to store and retrieve relevant historical sequences, seamlessly integrating them into the forecasting process. Our method dynamically combines global retrieval-based predictions with local model outputs, leading to superior performance. Extensive experiments validate the effectiveness of PFRP, demonstrating its ability to capture global correlations and enhance forecasting accuracy across different local prediction models.

\bibliography{aaai2026}


\clearpage
\newpage
\appendix

\section{Dataset Details}
\label{sec:dataset}
The Traffic, Electricity, Weather, and ETT datasets are widely used benchmarks for time series forecasting, covering various domains with diverse temporal patterns.

\begin{itemize}
    \item
\textbf{Traffic}\footnote{\url{https://pems.dot.ca.gov}} records hourly road occupancy rates collected from 862 sensors across San Francisco Bay Area freeways between 2015 and 2016. For univariate TSF, we use only the time series recorded by the last sensor. 
    \item
\textbf{Electricity}\footnote{\url{https://archive.ics.uci.edu/dataset/321/electricityloaddiagrams20112014}} contains hourly electricity consumption data from 321 customers from 2012 to 2014. In univariate TSF, we focus solely on the consumption series of the last customer. 
    \item 
\textbf{Weather}\footnote{\url{https://www.bgc-jena.mpg.de/wetter}} captures 21 meteorological factors monitored every 10 minutes by the Max Planck Biogeochemistry Institute in 2020. For univariate TSF, we select the carbon dioxide concentration series. 
    \item 
\textbf{ETT}\footnote{\url{https://github.com/zhouhaoyi/ETDataset}} consists of four subsets (ETTh1, ETTh2, ETTm1, and ETTm2) that record the oil temperature and load characteristics of two power transformers between 2016 and 2018, with hourly and 15-minute resolutions, respectively. For univariate TSF, we focus on the oil temperature series. 
\end{itemize}

As summarized in Table~\ref{tab:dataset}, these datasets provide a comprehensive evaluation framework for time series forecasting models, exhibiting diverse periodicities, seasonal patterns, and complex dependencies.

\begin{table}[ht]
  \centering
    \begin{tabular}{cccc}
    \toprule
    Dataset & Length & Frequency & Dataset split \\
    \midrule
    Traffic & 17544 & 1 Hour & 7:1:2 \\
    Electricity & 26304 & 1 Hour & 7:1:2\\
    Weather & 52696 & 10 Minutes & 7:1:2\\
    ETTh1 \& ETTh2 & 17420 & 1 Hour & 6:2:2\\
    ETTm1 \& ETTm2 & 69680 & 15 Minutes & 6:2:2\\
    \bottomrule
    \end{tabular}
      \caption{The statistics of each dataset. Length indicates the total number of time points. Frequency denotes the sampling interval of time points. Dataset split indicates the ratio of training, validation, and test sets.}
  \label{tab:dataset}%
\end{table}%

\section{Baseline Details}
\label{sec:baseline}
To assess the effectiveness of PFRP, we evaluate four state-of-the-art univariate forecasting models, each representing a distinct architectural paradigm: MLP-based, Transformer-based, and Convolution-based approaches. Specifically, we include \textbf{SparseTSF} (ICML 2024) and \textbf{DLinear} (AAAI 2023) from the MLP family, \textbf{PatchTST} (ICLR 2023) as a Transformer-based model, and \textbf{TimesNet} (ICLR 2023) as a Convolution-based model.  

\begin{itemize}
    \item \textbf{SparseTSF}\footnote{\url{https://github.com/lss-1138/SparseTSF}} is an ultra-lightweight forecasting model that downsamples time series into periodic subsequences and performs predictions independently on each subsequence, effectively transforming the original forecasting problem into a cross-period trend prediction task.
    \item \textbf{DLinear}\footnote{\url{https://github.com/cure-lab/LTSF-Linear}} introduces a decomposition-based linear framework that separately captures trend and seasonal components, achieving strong forecasting performance with a simple yet effective design.
    \item \textbf{PatchTST}\footnote{\url{https://github.com/yuqinie98/patchtst}} applies a patching mechanism, segmenting the time series into fixed-length patches before feeding them into a Transformer encoder, which enhances local feature extraction and long-range dependency modeling.
    \item \textbf{TimesNet}\footnote{\url{https://github.com/thuml/TimesNet}} converts one-dimensional time series into structured two-dimensional tensors, leveraging convolutional operations to model both intra-periodic and inter-periodic dependencies.
\end{itemize}

For a fair comparison, all experiments are conducted using our own implementations on the same hardware. We utilize official or publicly available codebases and adhere to the hyperparameter settings recommended in their respective papers.

\begin{figure*}[t]
\begin{center}
\includegraphics[width=0.98\linewidth]{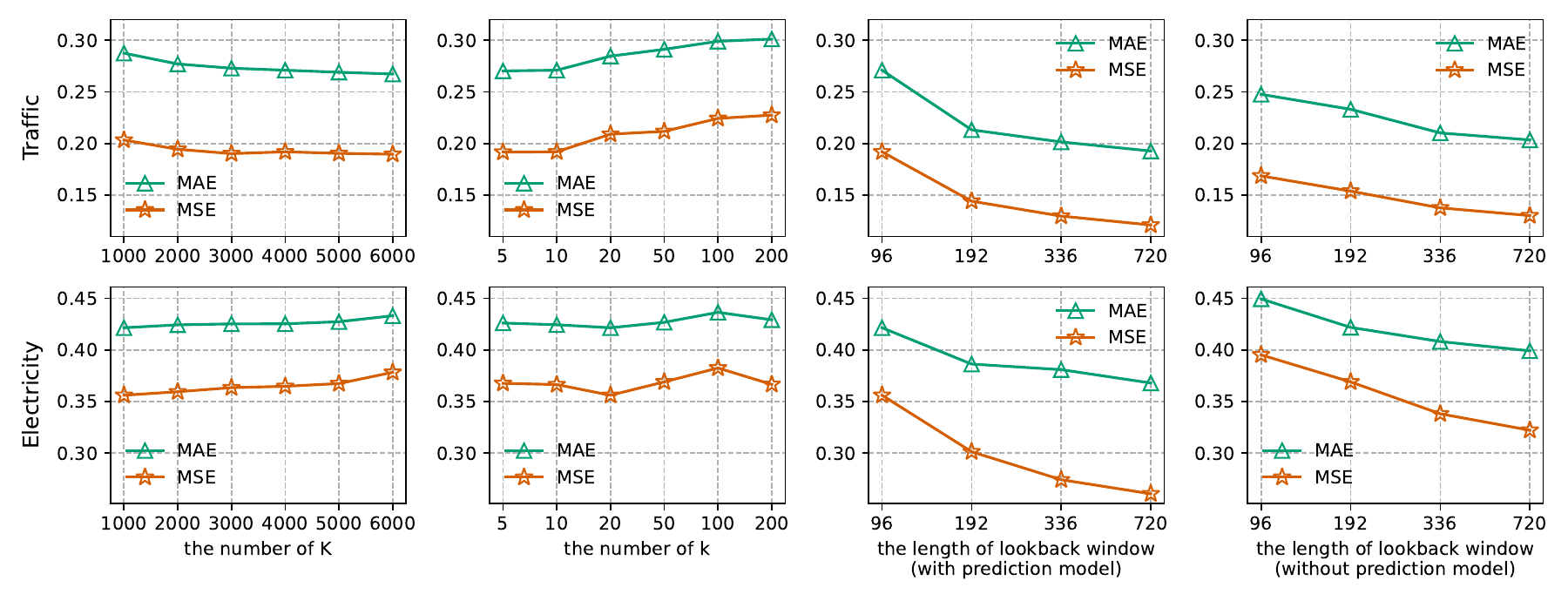}
\caption{The impact of key hyperparameters $K$ and $k$ in PFRP, as well as the lookback window length $L$, on forecasting performance. We explore the impact of $L$ on performance both with and without the local prediction model. See Figure~\ref{fig:sensitivity1} for the results on other five datasets.}
\label{fig:sensitivity}
\end{center}
\end{figure*}

\section{Hyperparameter Sensitivity Analysis}

In our PFRP, two key hyperparameters significantly influence performance: $K$, the number of historical samples retained in the GMB, and $k$, the number of relevant historical samples retrieved. A larger $K$ allows for storing a more diverse range of historical patterns but may also retain some rare patterns. Conversely, a smaller $K$ enables faster retrieval but may omit valuable historical information. Similarly, setting $k$ too large could result in retrieving irrelevant historical data, which might interfere with predictions, while a smaller $k$ limits the amount of historical information available. Therefore, careful selection of $K$ and $k$ is essential, as their optimal values often vary across datasets. As shown in Figure~\ref{fig:sensitivity}, for the Traffic dataset, larger $K$ values yield better results, while for the Electricity dataset, smaller $K$ values are more effective. Analysis of other five datasets are provided in the Tabel~\ref{tab:hyperparameter}. In addition, we investigate the impact of the lookback window length $L$ on model performance. To eliminate interference from the local prediction model, we also remove it to investigate the impact of $L$ on the global prediction performance of PFRP. Figure~\ref{fig:sensitivity} illustrates that longer lookback windows consistently lead to better results. This is because longer windows provide more information for local predictions and enable more accurate retrieval for global predictions, ultimately improving overall forecasting performance.

\begin{figure*}[t]
\centering
\includegraphics[width=0.98\linewidth]{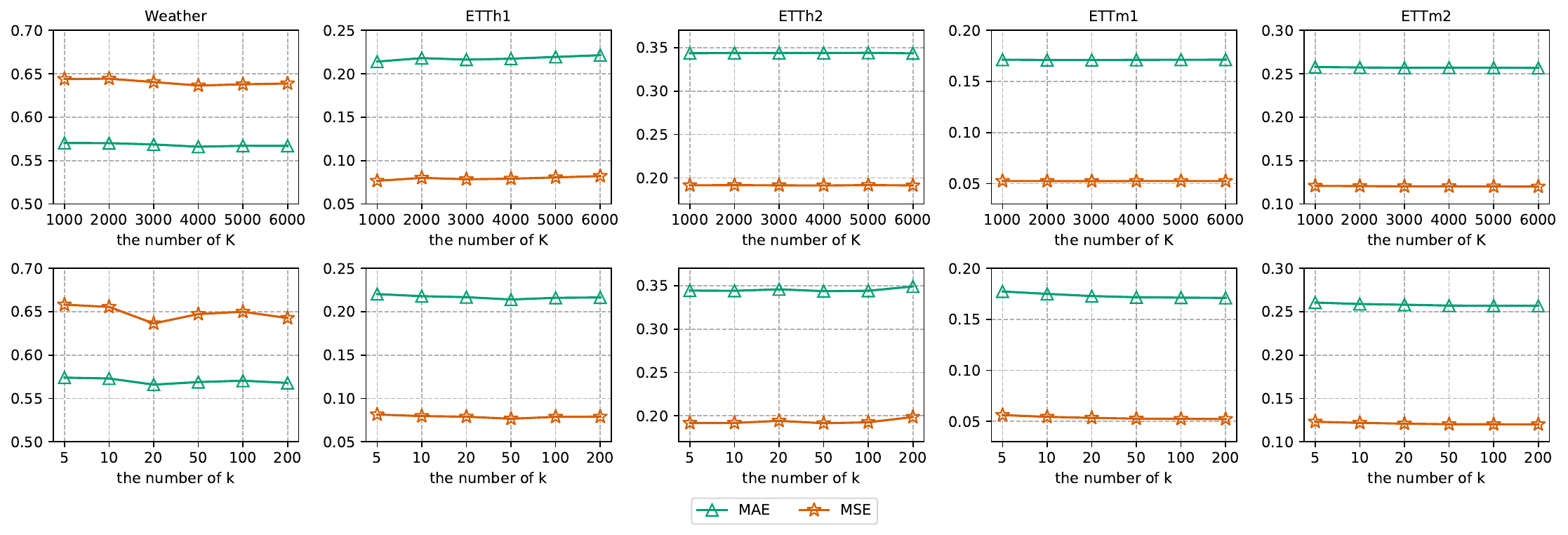}
\caption{Sensitivity analysis of hyperparameters $K$ and $k$ across different datasets.}
\label{fig:sensitivity1}
\end{figure*}

\begin{table}[ht]
\centering
\begin{tabular}{ccc}
    \toprule
    Dataset & $K$ & $k$ \\
    \midrule
    Traffic & 4000 & 10\\
    Electricity &1000 &20\\
    Weather&4000&20\\
    ETTh1&1000&50\\
    ETTh2&1000&50\\
    ETTm1&3000&200\\
    ETTm2&3000&100\\
    \bottomrule
\end{tabular}
\caption{Optimal hyperparameter configurations ($K$ and $k$) for each dataset.}
\label{tab:hyperparameter}
\end{table}

To further analyze the impact of hyperparameters $K$ and $k$ across five other datasets, we present their sensitivity curves in Figure~\ref{fig:sensitivity1}. These curves illustrate how varying $K$ and $k$ affects forecasting performance on each dataset. As observed, the optimal values of $K$ and $k$ vary across datasets due to differences in temporal patterns and historical dependencies. However, for four ETT datasets, the performance remains relatively stable across a range of $K$ and $k$ values, indicating a lower sensitivity to these hyperparameters. Additionally, Table~\ref{tab:hyperparameter} summarizes the best-performing configurations for each dataset. These configurations are determined through experimentation, striking a balance between retrieval effectiveness and computational efficiency.

\section{Visualization of Lookback Window Features}

\begin{figure}[ht]
\centering
\includegraphics[width=\columnwidth]{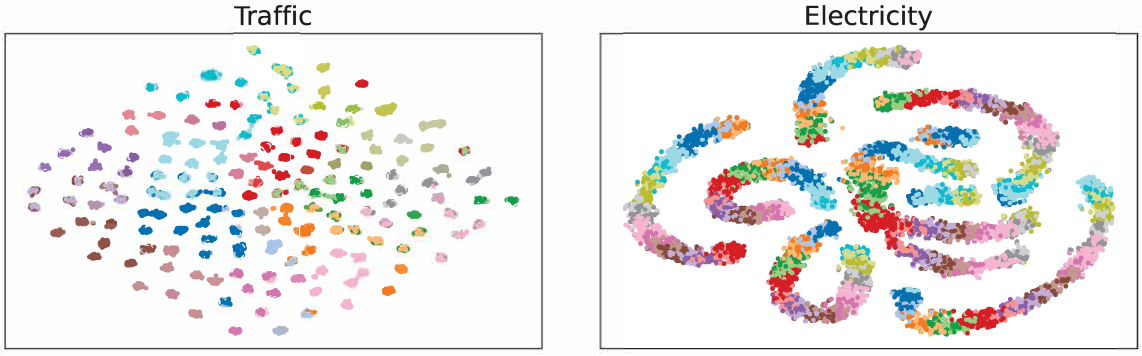}
\caption{T-SNE visualization of the lookback window features from all historical training samples on the Traffic and Electricity datasets.}
\label{fig:features}
\end{figure}

To evaluate whether PCL effectively learns meaningful lookback window features, we perform t-SNE visualization of the lookback window features from all historical training samples, as shown in Figure~\ref{fig:features}. Both the Traffic and Electricity datasets have an hourly granularity, with a minimum possible period of 24 identified using the autocorrelation function (ACF). Consequently, we assign the same color to lookback window features that are spaced 24 time steps apart. The visualization reveals that features with the same color are grouped into multiple smaller, tighter clusters, demonstrating that PCL effectively encodes the lookback windows by clustering features of similar sequences together.

\section{Periodicity Scores Calculation}

To quantify the periodicity of a univariate time series \( \{x_t\}_{t=1}^T \), we compute a normalized score in the range \([0, 1]\), combining autocorrelation at known periodic lags and entropy-based randomness.

\paragraph{Autocorrelation Function}
Let \( \mathcal{L} = \{l_1, l_2, \dots, l_K\} \) be a set of known periodic lags (e.g., 24 for daily, 168 for weekly). For each \( l_k \in \mathcal{L} \), we compute the autocorrelation:

\[
\text{ACF}(l_k) = \frac{\sum_{t=1}^{T-l_k} (x_t - \bar{x})(x_{t+l_k} - \bar{x})}{\sum_{t=1}^{T} (x_t - \bar{x})^2},
\]
where \( \bar{x} \) is the mean of the time series. The average autocorrelation score is then:

\[
\text{ACF}_{\text{score}} = \frac{1}{K} \sum_{k=1}^{K} \text{ACF}(l_k).
\]

\paragraph{Entropy Measure}
To assess randomness, we compute the normalized Shannon entropy. Let the time series values be discretized into \( B \) bins, and \( p_i \) be the probability of values falling into bin \( i \). Then the entropy is:

\[
H(x) = -\sum_{i=1}^{B} p_i \log p_i.
\]
Normalized by the maximum entropy \( \log B \), we have:
\[
\text{Entropy}_{\text{norm}} = \frac{H(x)}{\log B}.
\]
We define the inverse entropy as:
\[
\text{InvEntropy} = 1 - \text{Entropy}_{\text{norm}}.
\]

\paragraph{Final Periodicity Score}
Combining both metrics, the final periodicity score is defined as:
\[
\text{PeriodicityScore} = \text{ACF}_{\text{score}} \cdot \text{InvEntropy}.
\]
This score ranges from 0 (non-periodic) to 1 (highly periodic).

The periodicity scores of all the seven datasets are listed in Table~\ref{tab:periodicity}.

\begin{table}[ht]
\centering
\begin{tabular}{cc}
    \toprule
    Dataset & Periodicity score \\
    \midrule
    Traffic & 0.3202 \\
    Electricity & 0.1931 \\
    Weather& 0.1558\\
    ETTh1& 0.1281\\
    ETTm1& 0.1283\\
    ETTh2& 0.0727\\
    ETTm2& 0.0728\\
    \bottomrule
\end{tabular}
\caption{Periodicity score for each dataset.}
\label{tab:periodicity}
\end{table}

\begin{table*}[t]
  \centering
  \resizebox{0.99\linewidth}{!}{
        \begin{tabular}{c|c|cccc|cccc|cccc|cccc|c}
    \toprule
    \multicolumn{2}{c|}{Method} & \multicolumn{2}{c}{SparseTSF} & \multicolumn{2}{c|}{+PFRP} & \multicolumn{2}{c}{Dlinear} & \multicolumn{2}{c|}{+PFRP} & \multicolumn{2}{c}{PatchTST} & \multicolumn{2}{c|}{+PFRP} & \multicolumn{2}{c}{TimesNet} & \multicolumn{2}{c|}{+PFRP} & \multirow{2}[4]{*}{Impr.} \\
\cmidrule{1-18}    \multicolumn{2}{c|}{Metric} & MSE   & MAE   & MSE   & MAE   & MSE   & MAE   & MSE   & MAE   & MSE   & MAE   & MSE   & MAE   & MSE   & MAE   & MSE   & MAE   &  \\
    \midrule
    \multirow{5}[4]{*}{Traffic} & 96    & 0.2782  & 0.3463  & \textbf{0.1904 } & \textbf{0.2747 } & 0.3051  & 0.3993  & \textbf{0.1742 } & \textbf{0.2660 } & 0.1725  & 0.2623  & \textbf{0.1639 } & \textbf{0.2344 } & 0.2110  & 0.2995  & \textbf{0.1772 } & \textbf{0.2556 } & \multirow{5}[4]{*}{\textbf{17.4}} \\
          & 192   & 0.2274  & 0.2926  & \textbf{0.1742 } & \textbf{0.2539 } & 0.2570  & 0.3450  & \textbf{0.1701 } & \textbf{0.2598 } & 0.1750  & 0.2647  & \textbf{0.1617 } & \textbf{0.2324 } & 0.2015  & 0.2906  & \textbf{0.1727 } & \textbf{0.2547 } &  \\
          & 336   & 0.2164  & 0.2856  & \textbf{0.1673 } & \textbf{0.2510 } & 0.2518  & 0.3414  & \textbf{0.1709 } & \textbf{0.2653 } & 0.1721  & 0.2615  & \textbf{0.1634 } & \textbf{0.2367 } & 0.2067  & 0.3005  & \textbf{0.1749 } & \textbf{0.2618 } &  \\
          & 720   & 0.2397  & 0.3081  & \textbf{0.2357 } & \textbf{0.3042 } & 0.2972  & 0.3783  & \textbf{0.2018 } & \textbf{0.2970 } & 0.1991  & 0.2833  & \textbf{0.1961 } & \textbf{0.2630 } & 0.2468  & 0.3355  & \textbf{0.1948 } & \textbf{0.2803 } &  \\
\cmidrule{2-18}          & Avg.  & 0.2404  & 0.3082  & \textbf{0.1919 } & \textbf{0.2710 } & 0.2778  & 0.3660  & \textbf{0.1793 } & \textbf{0.2720 } & 0.1797  & 0.2680  & \textbf{0.1712 } & \textbf{0.2416 } & 0.2165  & 0.3065  & \textbf{0.1799 } & \textbf{0.2631 } &  \\
    \midrule
    \multirow{5}[4]{*}{Electricity} & 96    & 0.4813  & 0.5109  & \textbf{0.2923 } & \textbf{0.3792 } & 0.3898  & 0.4526  & \textbf{0.3295 } & \textbf{0.4109 } & 0.3245  & 0.4133  & \textbf{0.3090 } & \textbf{0.3995 } & 0.3773  & 0.4342  & \textbf{0.3347 } & \textbf{0.4148 } & \multirow{5}[4]{*}{\textbf{10.1}} \\
          & 192   & 0.4608  & 0.4988  & \textbf{0.3244 } & \textbf{0.3976 } & 0.3668  & 0.4366  & \textbf{0.3346 } & \textbf{0.4134 } & 0.3987  & 0.4435  & \textbf{0.3385 } & \textbf{0.4161 } & 0.3784  & 0.4345  & \textbf{0.3564 } & \textbf{0.4262 } &  \\
          & 336   & 0.4966  & 0.5172  & \textbf{0.3706 } & \textbf{0.4301 } & 0.3933  & 0.4543  & \textbf{0.3760 } & \textbf{0.4379 } & 0.4451  & 0.4695  & \textbf{0.3998 } & \textbf{0.4510 } & 0.4359  & 0.4661  & \textbf{0.4178 } & \textbf{0.4620 } &  \\
          & 720   & 0.5484  & 0.5560  & \textbf{0.4370 } & \textbf{0.4789 } & 0.4305  & 0.4881  & \textbf{0.4262 } & \textbf{0.4780 } & 0.5118  & 0.5214  & \textbf{0.5003 } & \textbf{0.5157 } & 0.5142  & 0.5228  & \textbf{0.4711 } & \textbf{0.5038 } &  \\
\cmidrule{2-18}          & Avg.  & 0.4968  & 0.5207  & \textbf{0.3561 } & \textbf{0.4214 } & 0.3951  & 0.4579  & \textbf{0.3666 } & \textbf{0.4351 } & 0.4200  & 0.4620  & \textbf{0.3869 } & \textbf{0.4456 } & 0.4264  & 0.4644  & \textbf{0.3950 } & \textbf{0.4517 } &  \\
    \midrule
    \multirow{5}[4]{*}{Weather} & 96    & 0.5006  & 0.4937  & \textbf{0.4785 } & \textbf{0.4808 } & 0.5617  & 0.5206  & \textbf{0.4749 } & \textbf{0.4801 } & \textbf{0.4699}  & \textbf{0.4741}  & 0.4782 & 0.4742 & 0.5164  & 0.4998  & \textbf{0.4769 } & \textbf{0.4798 } & \multirow{5}[4]{*}{\textbf{3.1}} \\
          & 192   & 0.5882  & 0.5457  & \textbf{0.5740 } & \textbf{0.5375 } & 0.6635  & 0.5736  & \textbf{0.6296 } & \textbf{0.5564 } & 0.6105  & 0.5514  & \textbf{0.5848 } & \textbf{0.5396 } & 0.6306  & 0.5630  & \textbf{0.5808 } & \textbf{0.5416 } &  \\
          & 336   & 0.6398  & 0.5766  & \textbf{0.6346 } & \textbf{0.5719 } & 0.7497  & 0.6082  & \textbf{0.7198 } & \textbf{0.5946 } & \textbf{0.6494}  & \textbf{0.5781}  & 0.6552 & 0.5800 & 0.6811  & 0.5918  & \textbf{0.6475 } & \textbf{0.5789 } &  \\
          & 720   & \textbf{0.8520}  & 0.6737  & 0.8587 & \textbf{0.6736 } & 0.9251  & 0.6820  & \textbf{0.8761 } & \textbf{0.6631 } & 0.8868  & 0.6842  & \textbf{0.8523 } & \textbf{0.6715 } & 0.8952  & 0.6865  & \textbf{0.8476 } & \textbf{0.6700 } &  \\
\cmidrule{2-18}          & Avg.  & 0.6452  & 0.5724  & \textbf{0.6365 } & \textbf{0.5659 } & 0.7250  & 0.5961  & \textbf{0.6751 } & \textbf{0.5735 } & 0.6542  & 0.5719  & \textbf{0.6426 } & \textbf{0.5663 } & 0.6808  & 0.5853  & \textbf{0.6382 } & \textbf{0.5676 } &  \\
    \midrule
    \multirow{5}[4]{*}{ETTh1} & 96    & 0.0598  & 0.1869  & \textbf{0.0574 } & \textbf{0.1829 } & 0.0644  & 0.1878  & \textbf{0.0638 } & \textbf{0.1893 } & \textbf{0.0563}  & \textbf{0.1783}  & 0.0575 & 0.1830 & \textbf{0.0565}  & \textbf{0.1826}  & 0.0569 & 0.1828 & \multirow{5}[4]{*}{\textbf{3.4}} \\
          & 192   & 0.0776  & 0.2147  & \textbf{0.0732 } & \textbf{0.2062 } & 0.0880  & 0.2217  & \textbf{0.0878 } & \textbf{0.2225 } & 0.0753  & 0.2102  & \textbf{0.0732 } & \textbf{0.2071 } & \textbf{0.0736}  & 0.2103  & 0.0737 & \textbf{0.2083 } &  \\
          & 336   & 0.0928  & 0.2403  & \textbf{0.0857 } & \textbf{0.2298 } & 0.1098  & 0.2567  & \textbf{0.1060 } & \textbf{0.2538 } & 0.0889  & 0.2335  & \textbf{0.0865 } & \textbf{0.2306 } & 0.0881  & 0.2322  & \textbf{0.0859 } & \textbf{0.2313 } &  \\
          & 720   & 0.1062  & 0.2569  & \textbf{0.0900 } & \textbf{0.2374 } & 0.2018  & 0.3715  & \textbf{0.1914 } & \textbf{0.3599 } & 0.1004  & 0.2501  & \textbf{0.0904 } & \textbf{0.2372 } & 0.0957  & 0.2438  & \textbf{0.0899 } & \textbf{0.2362 } &  \\
\cmidrule{2-18}          & Avg.  & 0.0841  & 0.2247  & \textbf{0.0766 } & \textbf{0.2141 } & 0.1160  & 0.2594  & \textbf{0.1122 } & \textbf{0.2564 } & 0.0802  & 0.2180  & \textbf{0.0769 } & \textbf{0.2145 } & 0.0785  & 0.2172  & \textbf{0.0766 } & \textbf{0.2147 } &  \\
    \midrule
    \multirow{5}[4]{*}{ETTh2} & 96    & \textbf{0.1370}  & 0.2865  & 0.1390 & \textbf{0.2850 } & \textbf{0.1346}  & \textbf{0.2820}  & 0.1348 & 0.2836 & 0.1408  & 0.2893  & \textbf{0.1360 } & \textbf{0.2820 } & 0.1353  & 0.2849  & \textbf{0.1323 } & \textbf{0.2807 } & \multirow{5}[4]{*}{\textbf{3.1}} \\
          & 192   & 0.1843  & 0.3354  & \textbf{0.1798 } & \textbf{0.3313 } & 0.1883  & 0.3353  & \textbf{0.1833 } & \textbf{0.3318 } & 0.1970  & 0.3520  & \textbf{0.1891 } & \textbf{0.3405 } & 0.1859  & 0.3386  & \textbf{0.1823 } & \textbf{0.3344 } &  \\
          & 336   & 0.2263  & 0.3798  & \textbf{0.2184 } & \textbf{0.3740 } & 0.2381  & 0.3851  & \textbf{0.2254 } & \textbf{0.3748 } & 0.2261  & 0.3787  & \textbf{0.2180 } & \textbf{0.3721 } & 0.2145  & 0.3704  & \textbf{0.2144 } & \textbf{0.3691 } &  \\
          & 720   & 0.2619  & 0.4130  & \textbf{0.2286 } & \textbf{0.3845 } & 0.3358  & 0.4751  & \textbf{0.3152 } & \textbf{0.4583 } & 0.2392  & 0.3925  & \textbf{0.2141 } & \textbf{0.3699 } & \textbf{0.2346}  & \textbf{0.3897}  & 0.2378 & 0.3924 &  \\
\cmidrule{2-18}          & Avg.  & 0.2024  & 0.3537  & \textbf{0.1915 } & \textbf{0.3437 } & 0.2242  & 0.3694  & \textbf{0.2147 } & \textbf{0.3621 } & 0.2008  & 0.3531  & \textbf{0.1893 } & \textbf{0.3411 } & 0.1926  & 0.3459  & \textbf{0.1917 } & \textbf{0.3442 } &  \\
    \midrule
    \multirow{5}[4]{*}{ETTm1} & 96    & 0.0310  & 0.1330  & \textbf{0.0286 } & \textbf{0.1254 } & 0.0322  & 0.1317  & \textbf{0.0318 } & \textbf{0.1315 } & 0.0294  & 0.1283  & \textbf{0.0288 } & \textbf{0.1260 } & 0.0305  & 0.1318  & \textbf{0.0288 } & \textbf{0.1262 } & \multirow{5}[4]{*}{\textbf{1.6}} \\
          & 192   & 0.0451  & 0.1626  & \textbf{0.0435 } & \textbf{0.1583 } & 0.0508  & 0.1678  & \textbf{0.0476 } & \textbf{0.1623 } & 0.0444  & 0.1599  & \textbf{0.0435 } & \textbf{0.1590 } & 0.0446  & 0.1608  & \textbf{0.0442 } & \textbf{0.1592 } &  \\
          & 336   & 0.0579  & 0.1864  & \textbf{0.0572 } & \textbf{0.1838 } & 0.0728  & 0.2008  & \textbf{0.0713 } & \textbf{0.1997 } & 0.0584  & 0.1854  & \textbf{0.0578 } & \textbf{0.1842 } & 0.0589  & 0.1865  & \textbf{0.0576 } & \textbf{0.1844 } &  \\
          & 720   & 0.0803  & 0.2188  & \textbf{0.0800 } & \textbf{0.2167 } & \textbf{0.0988}  & \textbf{0.2351}  & 0.1002 & 0.2375 & 0.0816  & 0.2205  & \textbf{0.0807 } & \textbf{0.2182 } & 0.0808  & 0.2183  & \textbf{0.0797 } & \textbf{0.2170 } &  \\
\cmidrule{2-18}          & Avg.  & 0.0536  & 0.1752  & \textbf{0.0523 } & \textbf{0.1710 } & 0.0637  & 0.1838  & \textbf{0.0627 } & \textbf{0.1828 } & 0.0534  & 0.1735  & \textbf{0.0527 } & \textbf{0.1719 } & 0.0537  & 0.1744  & \textbf{0.0526 } & \textbf{0.1717 } &  \\
    \midrule
    \multirow{5}[4]{*}{ETTm2} & 96    & 0.0785  & 0.2092  & \textbf{0.0671 } & \textbf{0.1866 } & 0.0720  & 0.1953  & \textbf{0.0693 } & \textbf{0.1913 } & 0.0692  & 0.1921  & \textbf{0.0665 } & \textbf{0.1842 } & 0.0686  & 0.1879  & \textbf{0.0675 } & \textbf{0.1870 } & \multirow{5}[4]{*}{\textbf{2.2}} \\
          & 192   & 0.1084  & 0.2489  & \textbf{0.1009 } & \textbf{0.2358 } & 0.1046  & 0.2399  & \textbf{0.1042 } & \textbf{0.2385 } & 0.1033  & 0.2381  & \textbf{0.1010 } & \textbf{0.2365 } & 0.1029  & 0.2385  & \textbf{0.1017 } & \textbf{0.2369 } &  \\
          & 336   & 0.1349  & 0.2820  & \textbf{0.1307 } & \textbf{0.2746 } & 0.1354  & 0.2793  & \textbf{0.1348 } & \textbf{0.2772 } & 0.1331  & 0.2771  & \textbf{0.1312 } & \textbf{0.2756 } & 0.1347  & 0.2795  & \textbf{0.1332 } & \textbf{0.2777 } &  \\
          & 720   & 0.1836  & 0.3337  & \textbf{0.1823 } & \textbf{0.3309 } & 0.1900  & 0.3343  & \textbf{0.1871 } & \textbf{0.3313 } & 0.1918  & 0.3411  & \textbf{0.1884 } & \textbf{0.3379 } & 0.1881  & 0.3365  & \textbf{0.1832 } & \textbf{0.3319 } &  \\
\cmidrule{2-18}          & Avg.  & 0.1263  & 0.2684  & \textbf{0.1203 } & \textbf{0.2570 } & 0.1255  & 0.2622  & \textbf{0.1239 } & \textbf{0.2596 } & 0.1244  & 0.2621  & \textbf{0.1218 } & \textbf{0.2585 } & 0.1236  & 0.2606  & \textbf{0.1214 } & \textbf{0.2584 } &  \\
    \midrule
    \multicolumn{2}{c|}{Impr.} & \multicolumn{4}{c|}{\textbf{8.4}} & \multicolumn{4}{c|}{\textbf{7.1}} & \multicolumn{4}{c|}{\textbf{3.5}} & \multicolumn{4}{c|}{\textbf{4.4}} &  \\
    \bottomrule
    \end{tabular}}
    \caption{Full univariate forecasting results for four baselines with and without PFRP. Lower metric values indicate better performance, and the better results are highlighted in bold. The lookback window length $L$ is set to 96. Avg. represents the average results across all prediction horizons $H= \{96, 192, 336, 720\}$. Impr. represents the improvement for each model and dataset.}
  \label{tab:full_result}%
\end{table*}%

\begin{table*}[t]
  \centering
  \resizebox{0.99\linewidth}{!}{
        \begin{tabular}{c|c|cccc|cccc|cccc|cccc}
    \toprule
    \multicolumn{2}{c|}{Method} & \multicolumn{2}{c}{SparseTSF} & \multicolumn{2}{c|}{+PFRP} & \multicolumn{2}{c}{Dlinear} & \multicolumn{2}{c|}{+PFRP} & \multicolumn{2}{c}{PatchTST} & \multicolumn{2}{c|}{+PFRP} & \multicolumn{2}{c}{TimesNet} & \multicolumn{2}{c}{+PFRP} \\
\cmidrule{1-18}    \multicolumn{2}{c|}{Metric} & MSE   & MAE   & MSE   & MAE   & MSE   & MAE   & MSE   & MAE   & MSE   & MAE   & MSE   & MAE   & MSE   & MAE   & MSE   & MAE  \\
    \midrule
    \multirow{4}[4]{*}{Traffic} & 96    & $\pm$0.000  & $\pm$0.000  & $\pm$0.001 & $\pm$0.000 & $\pm$0.001  & $\pm$0.001  & $\pm$0.001 & $\pm$0.001 & $\pm$0.001  & $\pm$0.001  & $\pm$0.001 & $\pm$0.001 & $\pm$0.001  & $\pm$0.001  & $\pm$0.000 & $\pm$0.001 \\
          & 192   & $\pm$0.001  & $\pm$0.001  & $\pm$0.001 & $\pm$0.001 & $\pm$0.001  & $\pm$0.001  & $\pm$0.001 & $\pm$0.001 & $\pm$0.001  & $\pm$0.001  & $\pm$0.001 & $\pm$0.001 & $\pm$0.000  & $\pm$0.000  & $\pm$0.001 & $\pm$0.001 \\
          & 336   & $\pm$0.001  & $\pm$0.001  & $\pm$0.001 & $\pm$0.001 & $\pm$0.001  & $\pm$0.001  & $\pm$0.004 & $\pm$0.003 & $\pm$0.001  & $\pm$0.001  & $\pm$0.001 & $\pm$0.001 & $\pm$0.002  & $\pm$0.001  & $\pm$0.001 & $\pm$0.001 \\
          & 720   & $\pm$0.001  & $\pm$0.002  & $\pm$0.001 & $\pm$0.001 & $\pm$0.001  & $\pm$0.001  & $\pm$0.002 & $\pm$0.001 & $\pm$0.003  & $\pm$0.001  & $\pm$0.001 & $\pm$0.001 & $\pm$0.001  & $\pm$0.003  & $\pm$0.001 & $\pm$0.001 \\
    \midrule
    \multirow{4}[4]{*}{Electricity} & 96   &   $\pm$0.001  & $\pm$0.001  & $\pm$0.001 & $\pm$0.001 & $\pm$0.001  & $\pm$0.000  & $\pm$0.000 & $\pm$0.001 & $\pm$0.001  & $\pm$0.001  & $\pm$0.001 & $\pm$0.001 & $\pm$0.000  & $\pm$0.001  & $\pm$0.001 & $\pm$0.000 \\
          & 192   & $\pm$0.002  & $\pm$0.002  & $\pm$0.001 & $\pm$0.001 & $\pm$0.001  & $\pm$0.001  & $\pm$0.001 & $\pm$0.003 & $\pm$0.004  & $\pm$0.001  & $\pm$0.001 & $\pm$0.001 & $\pm$0.003  & $\pm$0.001  & $\pm$0.001 & $\pm$0.001 \\
          & 336   & $\pm$0.002  & $\pm$0.004  & $\pm$0.003 & $\pm$0.003 & $\pm$0.001  & $\pm$0.004  & $\pm$0.001 & $\pm$0.001 & $\pm$0.004  & $\pm$0.002  & $\pm$0.001 & $\pm$0.001 & $\pm$0.003  & $\pm$0.001  & $\pm$0.002 & $\pm$0.003 \\
          & 720   & $\pm$0.003  & $\pm$0.002  & $\pm$0.004 & $\pm$0.004 & $\pm$0.005  & $\pm$0.002  & $\pm$0.004 & $\pm$0.004 & $\pm$0.003  & $\pm$0.005  & $\pm$0.001 & $\pm$0.003 & $\pm$0.006  & $\pm$0.006  & $\pm$0.003 & $\pm$0.002 \\
    \midrule
    \multirow{4}[4]{*}{Weather} & 96   &   $\pm$0.001  & $\pm$0.000  & $\pm$0.001 & $\pm$0.001 & $\pm$0.001  & $\pm$0.000  & $\pm$0.001 & $\pm$0.001 & $\pm$0.000  & $\pm$0.000  & $\pm$0.001 & $\pm$0.001 & $\pm$0.001  & $\pm$0.000  & $\pm$0.001 & $\pm$0.001 \\
          & 192   & $\pm$0.001  & $\pm$0.001  & $\pm$0.003 & $\pm$0.001 & $\pm$0.001  & $\pm$0.001  & $\pm$0.001 & $\pm$0.000 & $\pm$0.001  & $\pm$0.002  & $\pm$0.001 & $\pm$0.001 & $\pm$0.002  & $\pm$0.001  & $\pm$0.004 & $\pm$0.001 \\
          & 336   & $\pm$0.002  & $\pm$0.003  & $\pm$0.004 & $\pm$0.002 & $\pm$0.001  & $\pm$0.002  & $\pm$0.001 & $\pm$0.002 & $\pm$0.003  & $\pm$0.001  & $\pm$0.002 & $\pm$0.001 & $\pm$0.003  & $\pm$0.004  & $\pm$0.001 & $\pm$0.001 \\
          & 720   & $\pm$0.003  & $\pm$0.002  & $\pm$0.003 & $\pm$0.001 & $\pm$0.002  & $\pm$0.002  & $\pm$0.005 & $\pm$0.003 & $\pm$0.004  & $\pm$0.002  & $\pm$0.001 & $\pm$0.002 & $\pm$0.005  & $\pm$0.006 & $\pm$0.003 & $\pm$0.004 \\
    \midrule
    \multirow{4}[4]{*}{ETTh1} & 96  &    $\pm$0.001  & $\pm$0.001  & $\pm$0.001 & $\pm$0.001 & $\pm$0.001  & $\pm$0.001  & $\pm$0.001 & $\pm$0.002 & $\pm$0.002  & $\pm$0.001  & $\pm$0.001 & $\pm$0.001 & $\pm$0.001  & $\pm$0.001  & $\pm$0.001 & $\pm$0.001 \\
          & 192   & $\pm$0.003 & $\pm$0.002  & $\pm$0.001 & $\pm$0.001 & $\pm$0.001  & $\pm$0.001  & $\pm$0.004 & $\pm$0.002 & $\pm$0.005  & $\pm$0.001  & $\pm$0.001 & $\pm$0.001 & $\pm$0.002  & $\pm$0.003  & $\pm$0.003 & $\pm$0.003 \\
          & 336   & $\pm$0.004  & $\pm$0.002  & $\pm$0.001 & $\pm$0.001 & $\pm$0.001  & $\pm$0.001  & $\pm$0.005 & $\pm$0.002 & $\pm$0.001  & $\pm$0.001  & $\pm$0.001 & $\pm$0.001 & $\pm$0.001  & $\pm$0.004  & $\pm$0.002 & $\pm$0.003 \\
          & 720   & $\pm$0.003  & $\pm$0.002  & $\pm$0.004 & $\pm$0.005 & $\pm$0.001  & $\pm$0.003  & $\pm$0.002 & $\pm$0.003 & $\pm$0.007 & $\pm$0.006  & $\pm$0.005 & $\pm$0.008 & $\pm$0.005  & $\pm$0.004  & $\pm$0.002 & $\pm$0.004  \\
    \midrule
    \multirow{4}[4]{*}{ETTh2} & 96  &    $\pm$0.001  & $\pm$0.001  & $\pm$0.001 & $\pm$0.001 & $\pm$0.001  & $\pm$0.001  & $\pm$0.002 & $\pm$0.001 & $\pm$0.001  & $\pm$0.001 & $\pm$0.001 & $\pm$0.001 & $\pm$0.001  & $\pm$0.001  & $\pm$0.001 & $\pm$0.001 \\
          & 192   & $\pm$0.001  & $\pm$0.002  & $\pm$0.001 & $\pm$0.002 & $\pm$0.002 & $\pm$0.003 & $\pm$0.002 & $\pm$0.002 & $\pm$0.003  & $\pm$0.002  & $\pm$0.001 & $\pm$0.001 & $\pm$0.003 & $\pm$0.003 & $\pm$0.007 & $\pm$0.004 \\
          & 336   & $\pm$0.003  & $\pm$0.002  & $\pm$0.003 & $\pm$0.002 & $\pm$0.003  & $\pm$0.002  & $\pm$0.003 & $\pm$0.006 & $\pm$0.004  & $\pm$0.003  & $\pm$0.004 & $\pm$0.001 & $\pm$0.007  & $\pm$0.002  & $\pm$0.003 & $\pm$0.003 \\
          & 720   & $\pm$0.004  & $\pm$0.005 & $\pm$0.003 & $\pm$0.003 & $\pm$0.006 & $\pm$0.004  & $\pm$0.002 & $\pm$0.008 & $\pm$0.003  & $\pm$0.004  & $\pm$0.009 & $\pm$0.004 & $\pm$0.005  & $\pm$0.006 & $\pm$0.003 & $\pm$0.004  \\
    \midrule
    \multirow{4}[4]{*}{ETTm1} & 96   &   $\pm$0.002  & $\pm$0.001  & $\pm$0.002 & $\pm$0.001 & $\pm$0.001  & $\pm$0.002  & $\pm$0.001 & $\pm$0.003 & $\pm$0.001  & $\pm$0.001  & $\pm$0.001 & $\pm$0.001 & $\pm$0.003  & $\pm$0.004  & $\pm$0.003 & $\pm$0.002 \\
          & 192   & $\pm$0.002  & $\pm$0.002  & $\pm$0.001 & $\pm$0.001 & $\pm$0.001  & $\pm$0.001  & $\pm$0.006 & $\pm$0.001 & $\pm$0.001  & $\pm$0.001  & $\pm$0.001 & $\pm$0.001 & $\pm$0.003  & $\pm$0.004  & $\pm$0.005 & $\pm$0.002 \\
          & 336   & $\pm$0.004  & $\pm$0.001  & $\pm$0.001 & $\pm$0.001 & $\pm$0.001  & $\pm$0.007 & $\pm$0.001 & $\pm$0.002 & $\pm$0.001  & $\pm$0.001  & $\pm$0.001 & $\pm$0.005 & $\pm$0.004  & $\pm$0.003  & $\pm$0.003 & $\pm$0.002 \\
          & 720   & $\pm$0.005  & $\pm$0.003  & $\pm$0.002 & $\pm$0.001 & $\pm$0.001  & $\pm$0.003  & $\pm$0.005 & $\pm$0.002 & $\pm$0.007  & $\pm$0.005  & $\pm$0.008 & $\pm$0.005 & $\pm$0.004  & $\pm$0.003  & $\pm$0.002 & $\pm$0.004  \\
    \midrule
    \multirow{4}[4]{*}{ETTm2} & 96  &    $\pm$0.001  & $\pm$0.001  & $\pm$0.001 & $\pm$0.001 & $\pm$0.001  & $\pm$0.001  & $\pm$0.001 & $\pm$0.001 & $\pm$0.001  & $\pm$0.001  & $\pm$0.001 & $\pm$0.001 & $\pm$0.002  & $\pm$0.001  & $\pm$0.001 & $\pm$0.001 \\
          & 192   & $\pm$0.004  & $\pm$0.001  & $\pm$0.002 & $\pm$0.001 & $\pm$0.001  & $\pm$0.001  & $\pm$0.001 & $\pm$0.001 & $\pm$0.001  & $\pm$0.001  & $\pm$0.001 & $\pm$0.001 & $\pm$0.002  & $\pm$0.003  & $\pm$0.002 & $\pm$0.003 \\
          & 336   & $\pm$0.002  & $\pm$0.003  & $\pm$0.001 & $\pm$0.001 & $\pm$0.002  & $\pm$0.001  & $\pm$0.003 & $\pm$0.005 & $\pm$0.002  & $\pm$0.001  & $\pm$0.001 & $\pm$0.001 & $\pm$0.001  & $\pm$0.003  & $\pm$0.002 & $\pm$0.001 \\
          & 720   & $\pm$0.002  & $\pm$0.005  & $\pm$0.007 & $\pm$0.003 & $\pm$0.004  & $\pm$0.003  & $\pm$0.003 & $\pm$0.004 & $\pm$0.002  & $\pm$0.002 & $\pm$0.002 & $\pm$0.004 & $\pm$0.002  & $\pm$0.005  & $\pm$0.003 & $\pm$0.004 \\
    \bottomrule
    \end{tabular}}
    \caption{Standard deviations of three repeated experiments (with different random seeds) for all forecasting settings.}
  \label{tab:std}%
\end{table*}%

\begin{table*}[t]
  \centering
  \resizebox{0.99\linewidth}{!}{
  \begin{tabular}{c|c|cc|cccccc|cccc|cccc}
    \toprule
    \multicolumn{1}{c}{} &       & \multicolumn{2}{c|}{} & \multicolumn{6}{c|}{Retrieval Strategy}       & \multicolumn{4}{c|}{Encoder Type} & \multicolumn{4}{c}{Training Strategy} \\
\cmidrule{3-18}    \multicolumn{1}{r}{} &       & \multicolumn{2}{c|}{Ours} & \multicolumn{2}{c}{MSE} & \multicolumn{2}{c}{DTW} & \multicolumn{2}{c|}{PCC} & \multicolumn{2}{c}{PatchTST} & \multicolumn{2}{c|}{TimesNet} & \multicolumn{2}{c}{CL} & \multicolumn{2}{c}{PL} \\
\cmidrule{3-18}    \multicolumn{1}{r}{} &       & MSE   & MAE   & MSE   & MAE   & MSE   & MAE   & MSE   & MAE   & MSE   & MAE   & MSE   & MAE   & MSE   & MAE   & MSE   & MAE \\
    \midrule
    \multirow{5}[4]{*}{Traffic} & 96    & 0.1904  & 0.2747  & 0.2163  & 0.3028  & 0.2106  & 0.3010  & 0.2092  & 0.2973  & 0.1940  & 0.2740  & 0.1797  & 0.2643  & 0.2334  & 0.3068  & 0.2497  & 0.3381  \\
          & 192   & 0.1742  & 0.2539  & 0.1840  & 0.2658  & 0.1865  & 0.2680  & 0.1870  & 0.2660  & 0.2273  & 0.2930  & 0.1718  & 0.2541  & 0.1997  & 0.2718  & 0.2040  & 0.2802  \\
          & 336   & 0.1673  & 0.2510  & 0.1728  & 0.2543  & 0.1733  & 0.2558  & 0.1755  & 0.2571  & 0.1775  & 0.2612  & 0.1642  & 0.2505  & 0.1916  & 0.2665  & 0.2093  & 0.2825  \\
          & 720   & 0.2357  & 0.3042  & 0.1981  & 0.2777  & 0.1984  & 0.2782  & 0.2362  & 0.3033  & 0.1953  & 0.2758  & 0.1858  & 0.2705  & 0.2158  & 0.2862  & 0.2371  & 0.3040  \\
\cmidrule{2-18}          & Avg.  & 0.1919  & 0.2710  & 0.1928  & 0.2752  & 0.1922  & 0.2758  & 0.2020  & 0.2809  & 0.1985  & 0.2760  & 0.1754  & 0.2599  & 0.2101  & 0.2828  & 0.2250  & 0.3012  \\
    \midrule
    \multirow{5}[4]{*}{Electricity} & 96    & 0.2923  & 0.3792  & 0.2979  & 0.3893  & 0.3023  & 0.3853  & 0.3131  & 0.3986  & 0.2826  & 0.3747  & 0.3038  & 0.3877  & 0.3238  & 0.4005  & 0.3128  & 0.3938  \\
          & 192   & 0.3244  & 0.3976  & 0.3282  & 0.3962  & 0.3206  & 0.3929  & 0.3338  & 0.4000  & 0.3180  & 0.3911  & 0.3289  & 0.4008  & 0.3425  & 0.4045  & 0.3422  & 0.4065  \\
          & 336   & 0.3706  & 0.4301  & 0.3850  & 0.4314  & 0.3829  & 0.4297  & 0.3845  & 0.4308  & 0.3617  & 0.4222  & 0.3812  & 0.4340  & 0.3870  & 0.4340  & 0.3866  & 0.4329  \\
          & 720   & 0.4370  & 0.4789  & 0.4753  & 0.5015  & 0.4576  & 0.4911  & 0.4843  & 0.5062  & 0.4343  & 0.4820  & 0.4879  & 0.5050  & 0.4800  & 0.5009  & 0.4631  & 0.4969  \\
\cmidrule{2-18}          & Avg.  & 0.3561  & 0.4214  & 0.3716  & 0.4296  & 0.3659  & 0.4248  & 0.3789  & 0.4339  & 0.3492  & 0.4175  & 0.3754  & 0.4319  & 0.3833  & 0.4350  & 0.3762  & 0.4325  \\
    \bottomrule
    \end{tabular}
    }
    \caption{Full ablation results about GMB. We compare four retrieval strategies, three types of encoder, and three training strategies for training encoders. The column \textbf{Ours} represents that we adopt the retrieval strategy as \textbf{Feature Cosine}, the encoder type as \textbf{MLP}, and the training strategy as \textbf{PCL}. Lower metric values indicate better performance. The lookback window length $L$ is set to 96. Avg. represents the average results across all prediction horizons $H= \{96, 192, 336, 720\}$.}
  \label{tab:full_abla_GMB}%
\end{table*}%

\begin{table*}[t]
  \centering
  \resizebox{0.95\linewidth}{!}{
    \begin{tabular}{c|c|cc|cc|cc|cc|cc|cc}
    \toprule
    \multicolumn{2}{c|}{\multirow{2}[2]{*}{}} & \multicolumn{2}{c|}{SparseTSF} & \multicolumn{2}{c|}{SparseTSF+PFRP} & \multicolumn{2}{c|}{w/o confidence gate} & \multicolumn{2}{c|}{w/o output gate} & \multicolumn{2}{c|}{w/o both gates} & \multicolumn{2}{c}{w/o prediction model} \\
    \midrule
    \multicolumn{2}{c|}{} & MSE   & MAE   & MSE   & MAE   & MSE   & MAE   & MSE   & MAE   & MSE   & MAE   & MSE   & MAE \\
    \midrule
    \multirow{5}[4]{*}{Traffic} & 96    & 0.2782  & 0.3463  & 0.1904  & 0.2747  & 0.2724  & 0.3387  & 0.2125  & 0.2969  & 0.2131  & 0.2974  & \textbf{0.1662 } & \textbf{0.2431 } \\
          & 192   & 0.2274  & 0.2926  & 0.1742  & 0.2539  & 0.2267  & 0.2935  & 0.2071  & 0.2934  & 0.2074  & 0.2936  & \textbf{0.1636 } & \textbf{0.2408 } \\
          & 336   & 0.2164  & 0.2856  & 0.1673  & 0.2510  & 0.2153  & 0.2838  & 0.2060  & 0.2948  & 0.2043  & 0.2932  & \textbf{0.1611 } & \textbf{0.2427 } \\
          & 720   & 0.2397  & 0.3081  & 0.2357  & 0.3042  & 0.2397  & 0.3074  & 0.2265  & 0.3104  & 0.2265  & 0.3104  & \textbf{0.1833 } & \textbf{0.2637 } \\
\cmidrule{2-14}          & Avg.  & 0.2404  & 0.3082  & 0.1919  & 0.2710  & 0.2385  & 0.3058  & 0.2130  & 0.2989  & 0.2128  & 0.2987  & \textbf{0.1686 } & \textbf{0.2476 } \\
    \midrule
    \multirow{5}[4]{*}{Electricity} & 96    & 0.4813  & 0.5109  & \textbf{0.2923 } & \textbf{0.3792 } & 0.3116  & 0.4001  & 0.4818  & 0.5049  & 0.5632  & 0.5509  & 0.3108  & 0.3992  \\
          & 192   & 0.4608  & 0.4988  & \textbf{0.3244 } & \textbf{0.3976 } & 0.3567  & 0.4235  & 0.4837  & 0.5124  & 0.5434  & 0.5479  & 0.3551  & 0.4275  \\
          & 336   & 0.4966  & 0.5172  & \textbf{0.3706 } & \textbf{0.4301 } & 0.4159  & 0.4571  & 0.5235  & 0.5356  & 0.5798  & 0.5673  & 0.4161  & 0.4578  \\
          & 720   & 0.5484  & 0.5560  & \textbf{0.4370 } & \textbf{0.4789 } & 0.4997  & 0.5137  & 0.5672  & 0.5655  & 0.6187  & 0.5912  & 0.4988  & 0.5133  \\
\cmidrule{2-14}          & Avg.  & 0.4968  & 0.5207  & \textbf{0.3561 } & \textbf{0.4214 } & 0.3960  & 0.4486  & 0.5140  & 0.5296  & 0.5763  & 0.5643  & 0.3952  & 0.4494  \\
    \bottomrule
    \end{tabular}}
    \caption{Full ablation results about PFRP. We remove one or multiple of the key learnable components in PFRP. Lower metric values indicate better performance, and the better results are highlighted in bold. The lookback window length $L$ is set to 96. Avg. represents the average results across all prediction horizons $H= \{96, 192, 336, 720\}$.}
  \label{tab:full_abla_PFRP}%
\end{table*}%

\section{Full Results}

Tables~\ref{tab:full_result}, \ref{tab:full_abla_GMB}, and \ref{tab:full_abla_PFRP} provide the complete experimental results corresponding to Tables~\ref{tab:result}, \ref{tab:abla_GMB}, and \ref{tab:abla_PFRP} in the main paper, detailing all forecasting metrics across various prediction horizons. The standard deviations from three repeated experiments are reported in Table~\ref{tab:std}. We find that regardless of whether PFRP is added or not, the standard deviations are generally low, typically less than 0.005.

\section{More Visualization Results}

\begin{figure*}[t]
\centering
\includegraphics[width=0.99\linewidth]{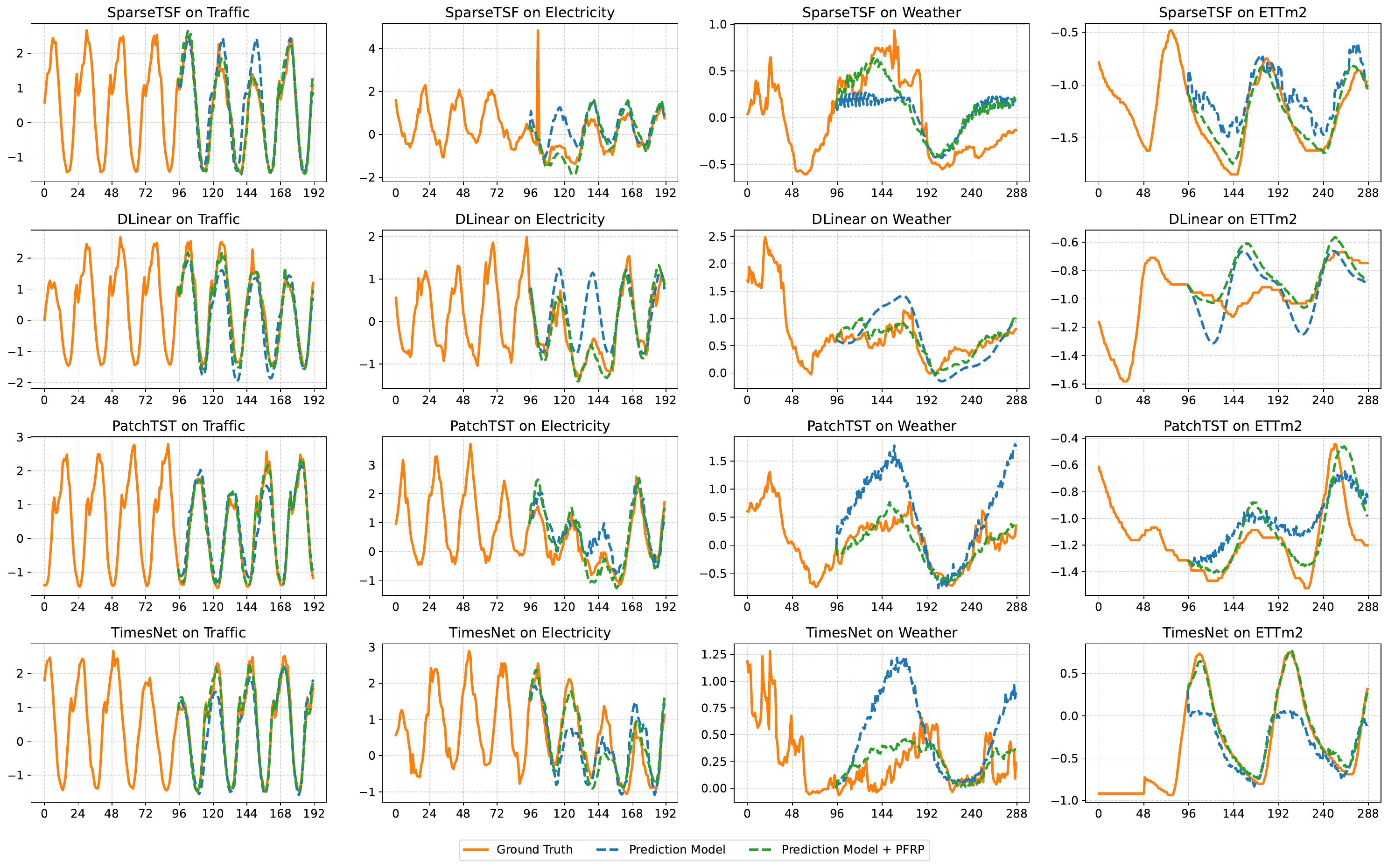}
\caption{Visualization of the prediction results for four baseline models, both with and without PFRP.}
\label{fig:result1}
\end{figure*}

In Figure~\ref{fig:result1}, we provide additional visual comparisons of baseline models with and without PFRP. Our observations indicate that across all major datasets, integrating PFRP consistently enhances the forecasting performance of each baseline model. These local prediction models often capture inaccurate periodic patterns, resulting in over- or underestimation of peaks. In contrast, PFRP effectively alleviates this issue by leveraging global historical information.

\section{Limitations and Future Work}
\label{sec:limitation}
Our method is specifically designed for univariate TSF. Extending it to multivariate TSF would require treating each variate as an independent univariate TSF task and constructing a separate GMB for each variate. However, this approach becomes computationally expensive when dealing with a large number of variates. Efficiently scaling our method to multivariate TSF remains an open challenge for future research. Additionally, timestamps, as an auxiliary feature, can help identify similar patterns. Developing strategies to integrate timestamp information into our method is another promising direction for future exploration.
\end{document}